\documentclass[10pt,twocolumn,letterpaper]{article}

\usepackage[pagenumbers]{cvpr}     % To produce the REVIEW version
\usepackage[table]{xcolor}
\definecolor{crieblue}{RGB}{225,237,248}
\usepackage{tabularx}
\usepackage{caption}
\usepackage{cuted}
\usepackage{multirow}
\usepackage{placeins}
\usepackage{bm}
\newtheorem{crietheorem}{Theorem}
\usepackage{cuted} %< for strip environment

\usepackage{currfile} %< for \currfilebase macro

\usepackage{caption} %< for "\captionof" commands (teaser)
\definecolor{cvprblue}{rgb}{0.21,0.49,0.74}
\usepackage[pagebackref,breaklinks,colorlinks,allcolors=cvprblue]{hyperref}

\def\confName{CVPR}
\def\confYear{2027}
\newcommand{\crietask}{complex redundant-instruction image editing}

\title{What Stays in the Picture? From Redundant Instructions to Dependency-Aware Image Editing}

\author{Zi-Jun Wang \quad Yu Lin \quad Zhilin Zhao\\
	Sun Yat-sen University\\
}

\begin{document}
\maketitle
\begin{strip}
  \centering
  \includegraphics[width=0.94\textwidth]{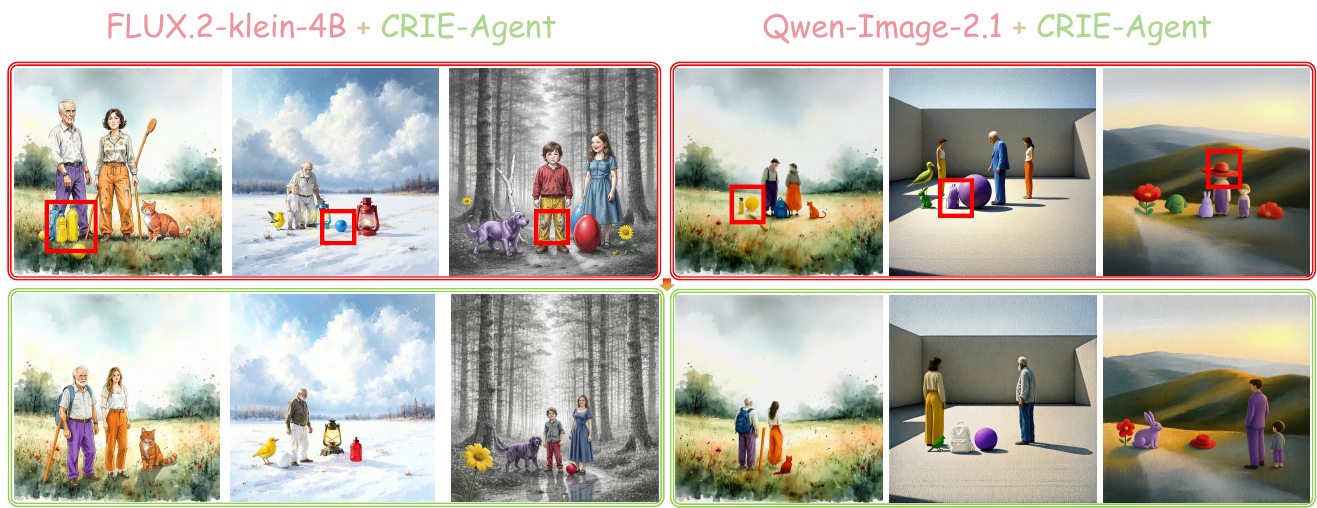}
  \captionsetup{hypcap=false}
  \captionof{figure}{Qualitative examples of \crietask \ with FLUX.2-klein-4B~\cite{flux2} and Qwen-Image-2.1~\cite{qwen2026image21}. Top: direct editing. Bottom: the corresponding editor augmented with CRIE-Agent. Red boxes highlight selected violations of the final editing requirements. CRIE-Agent satisfies the highlighted requirements in these examples.}
  \captionsetup{hypcap=true}
  \label{fig:tester}
\end{strip}
\begin{abstract}
We introduce complex redundant-instruction image editing, where a single request specifies multiple objects, their attributes, and spatial relations while revising or canceling earlier editing requirements.
Direct editing can fail to recover this intent, while splitting the request into successive steps can preserve obsolete edits and visual errors. 
Our key insight is to resolve the revision history into a final editing target that jointly guides planning and verification, so that obsolete requests do not become visual edits. 
Based on this insight, CRIE-Agent learns to consolidate revisions into object-level final intent and executes the retained objects in reference-first blocks, establishing referenced objects before their dependents. 
Verification then guides complete re-execution from the source image, with conditional layout hints, preventing visual errors from failed attempts from being carried into subsequent ones. 
We also construct CRIE-S, a benchmark containing 10,000 complex and redundant editing instructions with exact final intent annotations. 
Across five image-editing backends on CRIE-S, CRIE-Agent raises complete editing success from 0\% under direct editing to 27.5--50.0\%.
\end{abstract}    
\section{Introduction}
\label{sec:intro}

% P1 背景 + 问题
Image editing from natural language enables users to modify visual scenes through textual requests~\cite{anyedit,emuedit,instructpix2pix,flux1,qwenimage}. We focus on composing multiple foreground objects on a given background, where object attributes and spatial relations must be jointly specified. We study requests in which later clauses revise or cancel earlier requirements, such as changing an object's color or withdrawing a previously requested object. We formulate this setting as \crietask. The goal is to recover the final objects, attributes, and relations specified by the complete request and faithfully realize them in the source image.

\begin{figure}[t]
    \centering
    \includegraphics[width=1\linewidth]{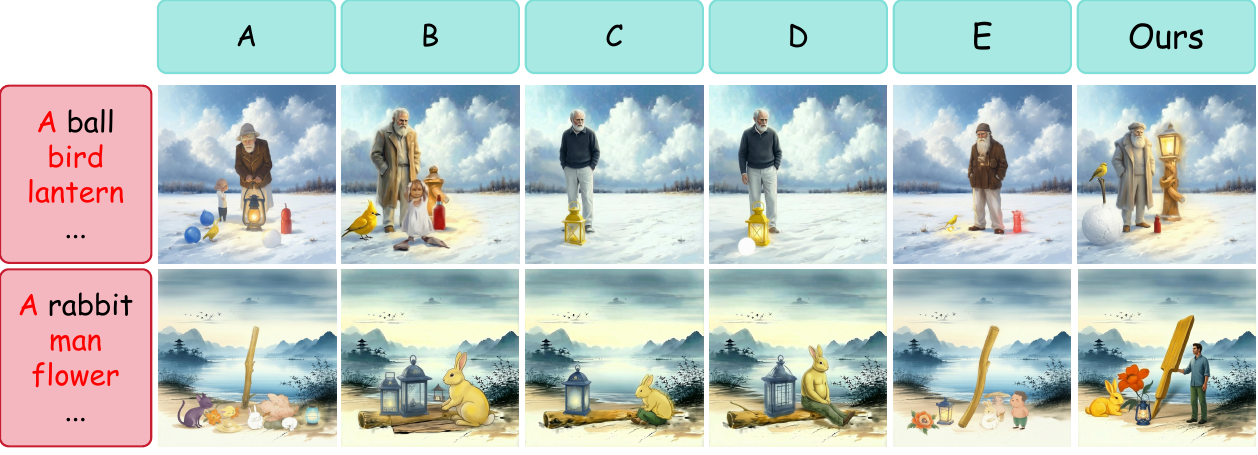}
    \caption{Qualitative comparison of instruction processing methods with Qwen-Image-Edit-2511. A: direct editing; B: punctuation-based splitting; C: VLM-based splitting; D: VLM reorganization into multiple editing prompts; E: VLM reorganization into a single prompt; Ours: our Intent Agent followed by single-step editing. The left column lists selected final requirements. The examples illustrate how intent recovery helps retain required objects and exclude superseded edits.}
    \label{fig:figure2}
\end{figure}

% P2 现有做法为何失败 + 任务分析。
Previous work has used multimodal language models to interpret editing requests and derive editing guidance~\cite{mgie,SmartEdit} or plan the execution of complex edits~\cite{xplanner,RePlan,GenArtist}.
Motivated by these strategies, Fig.~\ref{fig:figure2} compares direct editing with instruction-splitting and reorganization baselines.
Single-Step (A) sends the full request to the editor, while Punctuation split (B) and VLM split (C) divide it using punctuation marks and a general-purpose vision-language model, respectively.
VLM reorganization asks the model to express the final edits as either multiple simplified prompts (D) or a single combined prompt (E).
As the examples show, these approaches can still omit required objects, introduce additional ones, or produce incorrect attributes and spatial relations.

% P3 洞察 + 方法。
Even after these revisions are resolved, the editor must still satisfy multiple object attributes and spatial relations simultaneously while preserving the source image. 
Accordingly, our insight is to first resolve the revision history into a final editing target, then coordinate the execution of its interdependent requirements. This separates deciding which edits remain valid from planning how to realize them in the image.
Building on the planning and tool-use capabilities of LLM-based agents~\cite{yao2023react,shen2023hugginggpt,agashe2025agents,wang2025openhands}, we introduce CRIE-Agent, which couples two cooperating agents with an image editing model. 
The Intent Agent learns to resolve revisions in the instruction and recover the final objects, attributes, and spatial relations. 
The Execution Agent plans the editing order based on this recovered intent and invokes the image editing model to perform the edits. 
The recovered intent serves as a shared target for planning and verification, allowing the Execution Agent to assess whether the result satisfies the requirements and whether another editing attempt is needed. 

% P4 Benchmark + 结果。
To study this task, we construct CRIE-S, a benchmark containing 10,000 redundant editing instructions. 
For image editing evaluation, we pair the test instructions with source images selected from Style30K~\cite{li2024styletokenizer}, spanning diverse visual styles and scene types. 
We assess the edited images in terms of object correctness, attribute accuracy, spatial relations, and complete success. 
Complete success requires all object, attribute, and spatial relation checks to be answered and satisfied.
Experiments on CRIE-S show that CRIE-Agent improves the complete editing success rate from 0\% to 50.0\% with Qwen-Image-2.1~\cite{qwen2026image21}  and from 0\% to 27.5\% with FLUX.2-klein-4B~\cite{flux2}.
Our contributions are summarized as follows.
\begin{itemize}
    \item We formulate image editing with redundant instructions and construct CRIE-S, providing exact final intent annotations for studying intent recovery and image editing.
    \item We propose resolving the revision history into a final editing target before visual execution, so that obsolete requests do not become image edits. Following this principle, CRIE-Agent couples an Intent Agent that learns to recover this target with an Execution Agent that uses it to plan edits and verify their results.
    \item Experiments across five editing backends demonstrate consistent gains in complete instruction fulfillment, while controlled ablations establish the contributions of intent recovery and reference-first execution.
\end{itemize}
\section{Problem Formulation}
\label{sec:problem}

In this problem, a user provides a source image $I_0$ together with a single editing request $x$. The task involves both complexity and redundancy. Complexity arises from the need to jointly satisfy requirements for multiple objects, their attributes, and their relative positions and sizes. Redundancy arises from repeated revisions within the request, which may cancel or replace earlier requirements. We consider requests whose revisions can be consistently resolved into a unique final editing target, while realizing this target still requires satisfying the remaining requirements jointly.

\paragraph{Editing history.} We represent the editing history encoded in $x$ as an ordered sequence of atomic operations: 
\begin{equation}
E(x)=\left(e_1,e_2,\ldots,e_M\right),
\label{eq:edit_sequence}
\end{equation}
where each $e_m$ is an atomic edit, and the sequence follows the order in which the edits appear in the instruction. 
Applying these operations to $I_0$ under the predefined rules yields the final scene state $S^{*}$. 
Here, the scene state describes the task-relevant object states and relations, rather than a unique visual realization. 
The task is to produce an edited image $\hat{I}$ from $I_0$ and $x$ that satisfies this final target.

\paragraph{Object-level intent.} We describe the intended object states without retaining their revision history: 
\begin{equation}
Z=\left\{z_1,z_2,\ldots,z_N\right\},
\qquad
z_i=\left(o_i,\mathcal{A}_i\right),
\label{eq:intent_representation}
\end{equation}
where $o_i$ identifies an object and $\mathcal{A}_i$ contains its atomic attributes and relational constraints, such as color and position relative to another object. 

\paragraph{Final intent.} The final intent must preserve the semantics of the complete editing history while containing no removable object state or atomic attribute. 
Let $\mathcal{T}(I_0,E)$ denote the scene state obtained by applying an editing sequence $E$ to $I_0$, and let $\mathcal{S}(I_0,Z)$ denote the scene state specified by an intent $Z$ relative to $I_0$. 
We require the final intent $Z^{*}$ to satisfy semantic equivalence: 
\begin{equation}
\mathcal{S}(I_0,Z^{*})
=
\mathcal{T}\left(I_0,E(x)\right)
=
S^{*}.
\label{eq:semantic_equivalence}
\end{equation}
We further require $Z^{*}$ to be irreducible under deletion. 
For every valid representation $Z'$ obtained by removing an object state or an atomic attribute from $Z^{*}$, 
\begin{equation}
\mathcal{S}(I_0,Z')
\neq
\mathcal{S}(I_0,Z^{*}).
\label{eq:minimal_intent}
\end{equation}
Together, these conditions ensure that $Z^{*}$ retains the complete final semantics without unnecessary object states or attributes. 
\section{CRIE-S Benchmark}
\label{sec:crie-s}

CRIE-S provides matched redundant requests, atomic editing histories, and final-intent annotations for studying both revision resolution and the execution of complex editing requirements. Its construction samples object-level revision histories, extracts the retained final records, and verbalizes each history as a single request.

\paragraph{Redundancy in the instruction.}
We sample histories from operations that add, remove, move, recolor, resize, or change the pose of objects. An object may be added and later removed, a removed object may be added again, and an attribute may undergo several revisions. For a sampled history $E=(e_1,\ldots,e_M)$, let $S_0^{\mathrm{gen}}$ be the initially empty mapping of target objects to their attributes and $U_e^{\mathrm{gen}}$ its predefined update rule. Intermediate states satisfy
\[
S_m^{\mathrm{gen}}
=U_{e_m}^{\mathrm{gen}}(S_{m-1}^{\mathrm{gen}}),
\qquad m=1,\ldots,M.
\]
Objects are tracked by category name. Revisions update the corresponding fields, while removal deletes the object record and subsequent re-addition creates a new record without inheriting its previous attributes.

\paragraph{Complexity of the final edits.}
Annotations are derived from the final states of the retained objects. Let $V^*$ be their identifiers in $S_M^{\mathrm{gen}}$, and let $\mathcal A_o^*$ be the final attribute and relation record extracted for $o$. The retained-object set satisfies $|V^*|=5$, and the final intent is
\[
Z^*=\{(o,\mathcal A_o^*):o\in V^*\}.
\]
The five-object constraint applies to the retained target records. The numbers of intermediate objects and revisions are not fixed. Each sample stores its instruction, underlying atomic history, and final object records. Even after the revisions are resolved, editing requires jointly satisfying the attributes of multiple objects and their relative positions and sizes. These cross-object requirements make visual execution challenging beyond recovering the final intent.

Position and relative-size descriptions form a joint graph of object references. A directed dependency runs from a referenced object to the object whose position or size description uses it. Position relations include \emph{left of} and \emph{right of}. Samples whose final reference graphs contain cycles are rejected and regenerated. For valid samples, a topological ordering arranges the final descriptions so that every referenced object precedes its dependents.

Each atomic operation is expressed using a manually designed template. Colloquial connecting phrases join the clauses in their original order, producing one request while preserving its revision history. CRIE-S contains 10,000 samples, generated using a fixed random seed and shuffled using a separate seed before being split into 9,000 training, 800 validation, and 200 test examples. 
The splits share instruction templates and object and attribute vocabularies, with overlap in object-category combinations. Complete editing requests and final-intent records have no duplicates within or across splits. 
Further split details are provided in Appendix~\ref{sec:dataset_split}.
\begin{figure*}[t]
    \centering
    \includegraphics[width=0.95\linewidth]{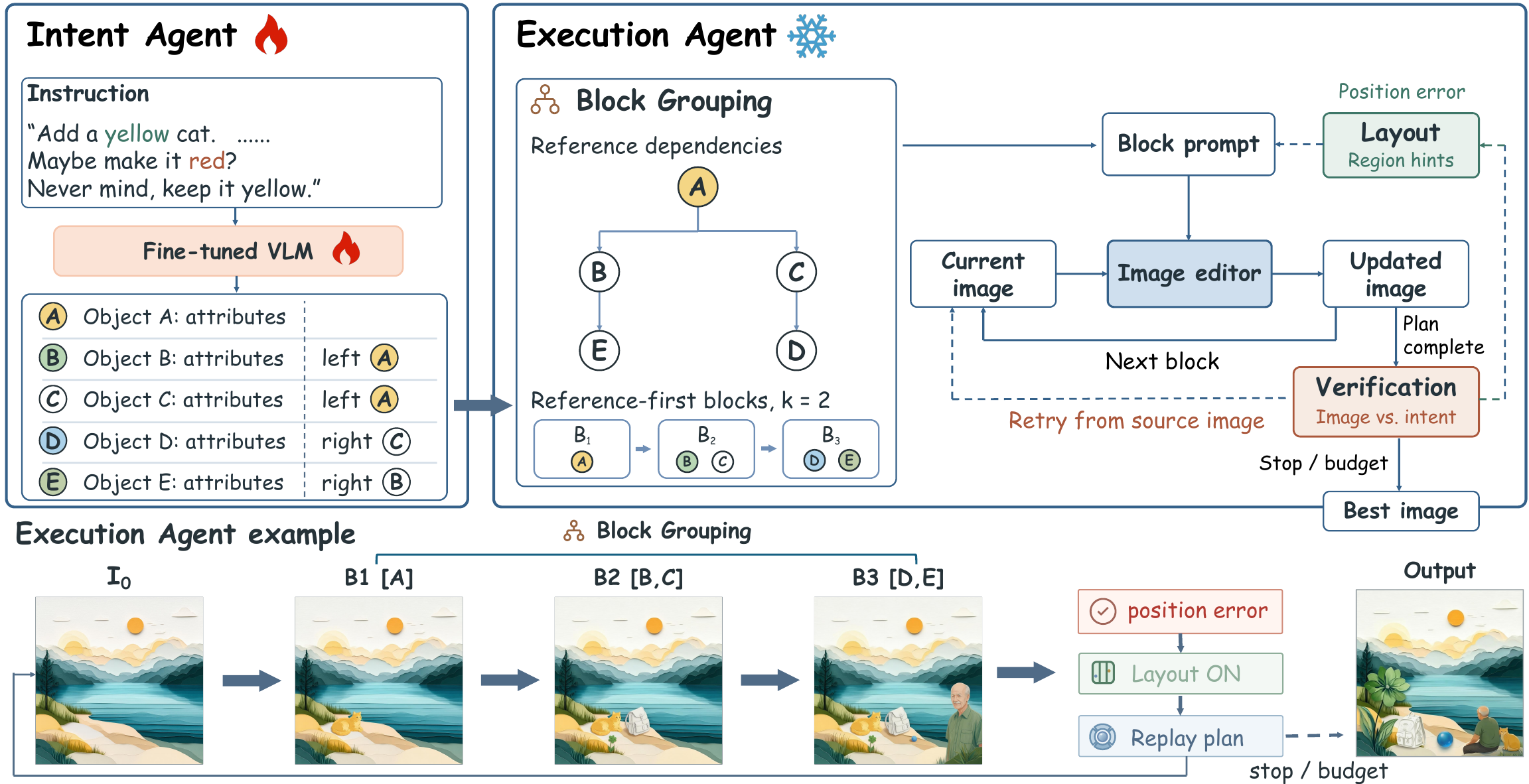}
    \caption{Overview of CRIE-Agent. A LoRA-tuned Intent Agent resolves instruction revisions into final object records. The Execution Agent groups these records into reference-first blocks and edits them sequentially. Verification controls stopping, candidate selection, and re-execution from the source image; Layout provides horizontal hints for eligible retries. The bottom example shows $K=2$ grouping and a layout-guided retry.}
    \label{fig:method}
\end{figure*}

\section{CRIE-Agent}
\label{sec:method}

CRIE-Agent separates recovering the final editing requirements from realizing them in the image. Resolving the revision history identifies which requirements remain valid, while visual execution must still satisfy multiple objects, attributes, and relations jointly. The Intent Agent produces final object records, and the Execution Agent plans their execution, edits the source image, and uses verification feedback to guide retries and select an output. Figure~\ref{fig:method} summarizes the overall framework.

\subsection{Intent Agent}
\label{sec:intent_agent}

Given the complete instruction $x$, the Intent Agent predicts records $\hat Z$ of the retained objects and their final attributes and relations. It resolves withdrawals and revisions before editing, providing a common target for planning and verification.

We fine-tune Qwen3-VL-8B-Instruct~\cite{qwen3vl2025} on CRIE-S using LoRA~\cite{lora}. Let $\theta_0$ denote the frozen pretrained parameters and $\phi$ the trainable LoRA parameters. The annotated intent $Z^*$, represented in Eq.~\eqref{eq:intent_representation}, is serialized as a JSON token sequence $y=(y_1,\ldots,y_L)$. Its conditional token likelihood factorizes as
\[
P_{\theta_0,\phi}(y_{1:L}\mid x)
=\prod_{\ell=1}^{L}
p_{\theta_0,\phi}(y_\ell\mid x,y_{<\ell}).
\]
Training uses target-token cross-entropy, with the sequence negative log-likelihood given by
\begin{equation}
\mathcal L_{\mathrm{intent}}
=-\sum_{\ell=1}^{L}
\log p_{\theta_0,\phi}(y_\ell\mid x,y_{<\ell}).
\label{eq:intent_loss}
\end{equation}
Only $\phi$ is optimized, while $\theta_0$ remains fixed. Training settings, the input template, and the JSON output format are provided in Appendix~\ref{sec:model_settings}.

\subsection{Execution Agent}
\label{sec:execution_agent}

Recovering the final intent removes ambiguity from the revision history, but the resulting editing task remains complex.
The Execution Agent uses Block Grouping to construct a fixed editing plan from the recovered object records. 
It executes the complete plan from the source image and verifies the resulting candidate. 
Feedback from verification controls retries and Layout activation before re-execution. 
The output is selected from the completed candidates using the recorded checks.

\subsubsection{Block Grouping}
\label{sec:block_grouping}

Block Grouping organizes objects into editing blocks, determining which objects are handled together and in what order. We adopt reference-first scheduling: objects used as references are edited before those whose positions or sizes depend on them.

\paragraph{Reference graph construction.}
We construct a reference graph from the position and relative-size descriptions in the recovered intent $\hat Z$. Let $V$ contain one node for each recovered object $o_v$. The graph is defined as
\begin{equation}
\begin{aligned}
G_{\mathrm{ref}}&=(V,E_{\mathrm{ref}}),\\
E_{\mathrm{ref}}&=\{(u,v):o_v\text{ uses }o_u\text{ as a reference}\}.
\end{aligned}
\label{eq:relation_graph}
\end{equation}
Each edge points from a referenced object to the object whose position or relative size depends on it. These dependencies determine the editing order, while the original position and size requirements are retained in the editing prompts.

\paragraph{Dependency layers and editing blocks.}
We first arrange the objects into dependency layers, then complete every layer before proceeding to the next. Within a layer, consecutive objects from the recovered record order are packed into bounded blocks.

Assume $V\ne\varnothing$, $K$ is a positive integer, and $G_{\mathrm{ref}}$ is acyclic. Write $\operatorname{Pred}(v)=\{u:(u,v)\in E_{\mathrm{ref}}\}$. Define reference depth along a topological order:
\[
d(v)=
\begin{cases}
1,&\operatorname{Pred}(v)=\varnothing,\\
1+\max_{u\in\operatorname{Pred}(v)}d(u),&\text{otherwise}.
\end{cases}
\]
The layers are $L_h=\{v:d(v)=h\}$, with $H=\max_{v\in V}d(v)$. Within each layer, objects retain their order in the predicted JSON array and form full blocks of size $K$, followed by at most one nonempty smaller block. The plan satisfies
\begin{equation}
\begin{gathered}
\Pi=(B_1,\ldots,B_T),\qquad
V=\biguplus_{t=1}^{T}B_t,\\
1\le |B_t|\le K,\qquad
(u,v)\in E_{\mathrm{ref}}\Rightarrow b(u)<b(v),
\end{gathered}
\label{eq:block_partition}
\end{equation}
where $b(v)$ is the index of the block containing $v$. Thus every referenced object is scheduled in an earlier block than its dependents. Each block corresponds to one editor call, and the plan remains fixed across attempts. In Appendix~\ref{sec:theory}, Theorem~\ref{thm:reference_feasibility} establishes that this construction schedules every object exactly once, preserves the recorded reference dependencies, and satisfies the block capacity. Theorem~\ref{thm:call_efficiency} bounds the number of editor calls used by one complete execution of the plan.

\subsubsection{Layout}
\label{sec:layout}

Layout adds horizontal guidance for a retry when the preceding candidate is judged to have no missing target objects but has a position error. It supplies region hints that supplement the original relation descriptions while retaining the fixed block plan.

For attempt $r$, let $C_r$ denote the completed candidate and $\lambda_r$ its layout guidance, with $\lambda_1=\varnothing$. The feedback from $C_r$ controls Layout activation for the retry. In the five-object setting, the existing deterministic order consistent with the predicted left/right relations assigns objects to far-left, left, center, right, and far-right regions. These hints accompany the prompts when the complete plan is rerun from $I_0$.

\subsubsection{Verification and Re-execution}
\label{sec:verification}

Verification and re-execution form a feedback loop for retry control and candidate selection. The verifier checks the result against the recovered requirements. The controller then stops or repeats the complete plan and finally selects a candidate using the recorded checks.

\paragraph{Complete attempts.}
A complete attempt executes the fixed block plan from the source image. Its prompts combine the object descriptions and relations in each block with the available layout guidance and include instructions to preserve the background and visual style.

Write $r$ for the attempt index and $t$ for the block index. The prompt $p(B_t,\lambda_r)$ includes region hints when Layout is activated. Execution is
\begin{equation}
\begin{aligned}
I_{r,0}&=I_0,\\
I_{r,t}&=\mathcal F\bigl(I_{r,t-1},p(B_t,\lambda_r)\bigr),
\quad t=1,\ldots,T.
\end{aligned}
\label{eq:execute}
\end{equation}
The complete candidate is $C_r=I_{r,T}$. Every retry restarts the same plan from $I_0$.

\paragraph{Checks and control.}
The verifier records object, attribute, spatial-relation, and visual-defect checks. Detected object and position errors determine whether to repeat the complete plan.

Write the check results as $\mathbf v_r=\mathcal V(C_r,\hat Z)$, and let $m_r,e_r,h_r\in\{0,1\}$ indicate a detected missing target object, extra object, and position error, respectively. The retry trigger is $\tau_r=\mathbf{1}\{m_r\vee e_r\vee h_r\}$. The controller stops when $\tau_r=0$ or after three attempts. For the $1\le R\le3$ completed candidates, the selected output is $\hat I=\mathsf{Select}((C_r,\mathbf v_r)_{r=1}^{R})$. Selection prioritizes object correctness, then attribute and relation correctness, and finally visual quality. Appendix~\ref{sec:verification_details} summarizes the controller.

\begin{figure*}[t]
\centering
\includegraphics[width=0.95\textwidth]{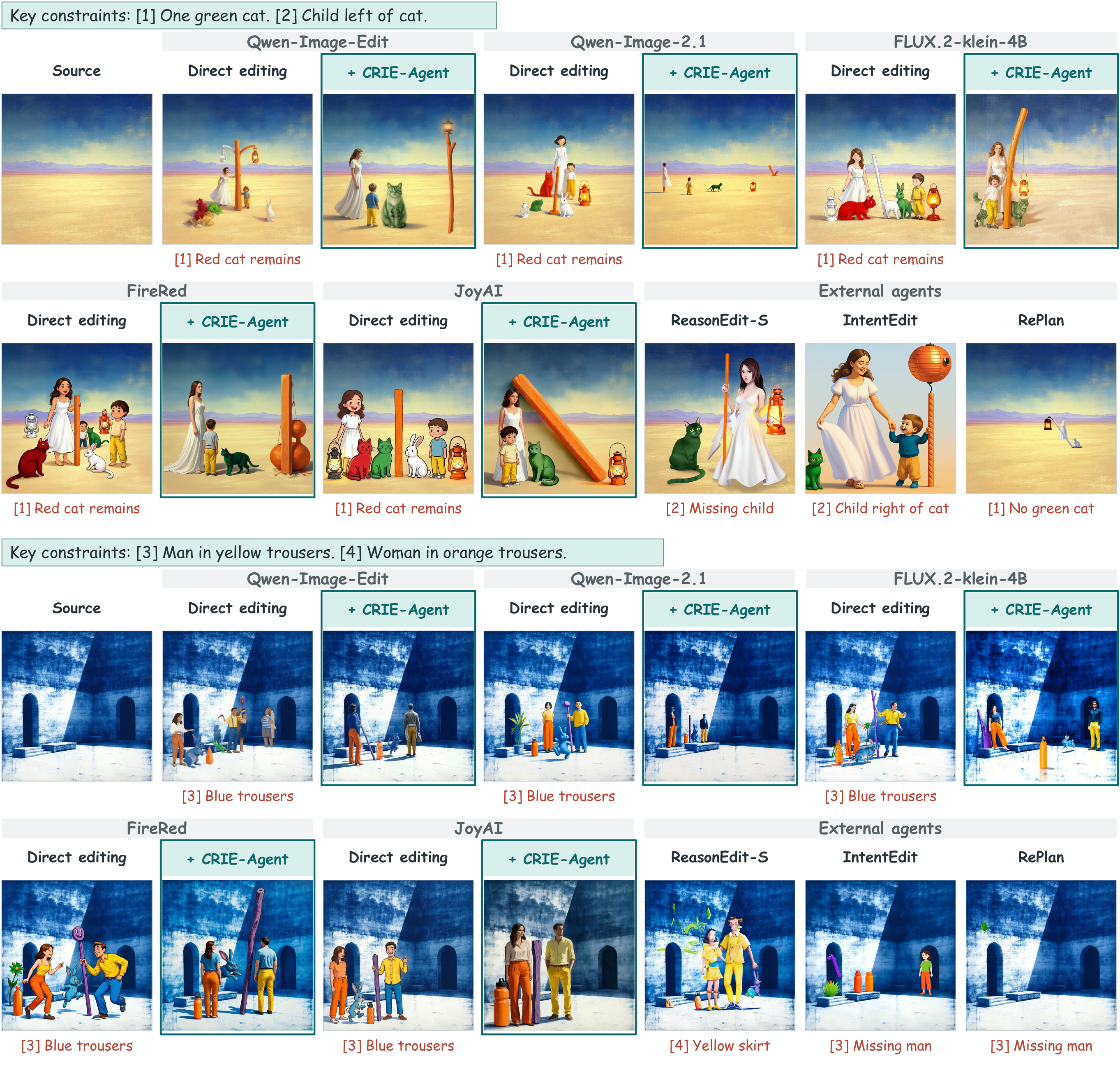}
\caption{Qualitative comparison of direct editing, existing editing agents, and CRIE-Agent on two CRIE-S examples. For each base editor, direct editing and CRIE-Agent outputs are paired. Numbered annotations identify violations of the key requirements listed above each example.}
\label{fig:qualitative}
\vspace{-2mm}
\end{figure*}

\begin{table}[t]
\centering
\caption{Comparison of base image editors and editing agents on the CRIE-S test set. O., A., R. and S. denote object correctness, attribute accuracy, relation accuracy and complete success.}
\label{tab:main}
\begingroup
\small
\renewcommand{\arraystretch}{1.08}
\begin{tabular}{@{}lccccc@{}}
\toprule
Method & O. $\uparrow$ & A. $\uparrow$ & R. $\uparrow$ & S. $\uparrow$ \\
\midrule
ReasonEdit-S & 55.2 & 60.1 & 35.6  & 0.0 \\
IntentEdit & 71.0 & 70.6 & 42.8  & 1.0 \\
RePlan & 39.4 & 18.1 & 7.8 &  0.0 \\
\midrule
Qwen-Image-Edit 2511 & 54.4 & 69.1 & 57.1  & 0.0 \\
\rowcolor{crieblue}
\quad \textbf{+ CRIE-Agent (Ours)} & \textbf{94.6} & \textbf{92.0} & \textbf{87.5}  & \textbf{39.5} \\
\midrule
Qwen-Image-2.1 & 57.9 & 83.6 & 69.9  & 0.0 \\
\rowcolor{crieblue}
\quad \textbf{+ CRIE-Agent (Ours)} & \textbf{96.8} & \textbf{93.5} & \textbf{88.0} & \textbf{50.0} \\
\midrule
FLUX.2-klein-4B & 54.1 & 76.4 & 65.4  & 0.0 \\
\rowcolor{crieblue}
\quad \textbf{+ CRIE-Agent (Ours)} & \textbf{91.8} & \textbf{92.0} & \textbf{85.8}  & \textbf{27.5} \\
\midrule
FireRed-Image-Edit-1.1 & 52.7 & 83.0 & 72.9  & 0.0 \\
\rowcolor{crieblue}
\quad \textbf{+ CRIE-Agent (Ours)} & \textbf{94.4} & \textbf{95.0} & \textbf{86.1}  & \textbf{36.5} \\
\midrule
JoyAI-Image-Edit & 61.5 & 84.4 & 80.0  & 0.0 \\
\rowcolor{crieblue}
\quad \textbf{+ CRIE-Agent (Ours)} & \textbf{97.0} & \textbf{93.4} & \textbf{88.1}  & \textbf{46.0} \\
\bottomrule
\end{tabular}
\endgroup
\end{table}

\begin{table}[t]
\centering
\caption{Comparison of instruction processing methods on the CRIE-S test set using Qwen-Image-Edit 2511. Strategy indicates single-step or multi-step execution. Metrics follow Table~\ref{tab:main}.}
\label{tab:abl_intent}
\begingroup
\small
\setlength{\tabcolsep}{5pt}
\renewcommand{\arraystretch}{1.10}
\begin{tabular}{@{}llccccc@{}}
\toprule
Method & Strategy & O. $\uparrow$ & A. $\uparrow$ & R. $\uparrow$ & S. $\uparrow$ \\
\midrule
Single-Step & Single-step & 54.4 & 69.1 & 57.1 & 0.0 \\
\midrule
Punctuation split & Multi-step & 68.4 & 60.7 & 40.8 & 1.0 \\
\midrule
VLM split & Multi-step & 68.3 & 62.2 & 40.3 & 1.0 \\
\midrule
\multirow{2}{*}{VLM reorganization}
& Multi-step & 70.3 & 62.9 & 41.9 & 3.5 \\
& Single-step & 79.2 & 78.8 & 58.8 & 5.5 \\
\midrule
\rowcolor{crieblue}
\textbf{Intent Agent} & Single-step & \textbf{92.0} & \textbf{88.0} & \textbf{69.4} & \textbf{11.0} \\
\bottomrule
\end{tabular}
\endgroup
\end{table}
\begin{figure*}[t]
\centering
\includegraphics[width=0.97\textwidth]{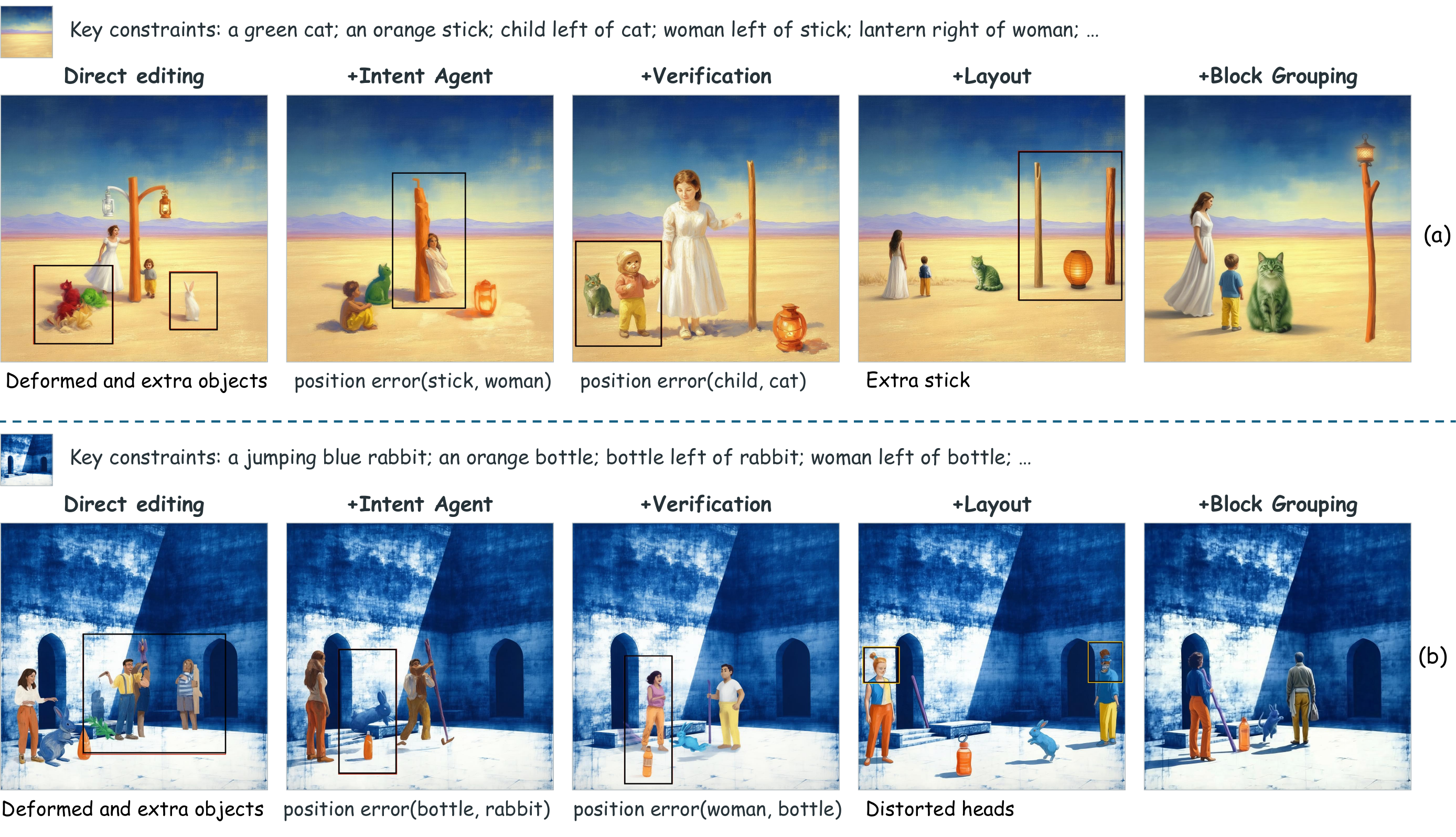}
\caption{Qualitative ablation of CRIE-Agent with Qwen-Image-Edit 2511. From left to right: direct editing, Intent Agent, and the successive addition of verification, layout, and block grouping. Boxes highlight object and spatial errors.}
\label{fig:execution_qualitative}
\vspace{-2mm}
\end{figure*}

\section{Experiments}
\label{sec:experiments}

\subsection{Evaluation Protocol}

\paragraph{Data.}
CRIE-S contains 10,000 instructions, split into 9,000 training, 800 validation, and 200 test samples. We train the Intent Agent on the training split and evaluate image editing on all test instructions, paired with source images from Style30K~\cite{li2024styletokenizer}. We additionally evaluate on CompBench's multi-object addition subset using two Qwen editors (Appendix~\ref{sec:compbench}).

\paragraph{Compared methods.}
We evaluate five base editors: Qwen-Image-Edit~\cite{qwenimage}, Qwen-Image-2.1~\cite{qwen2026image21}, FLUX.2-klein-4B~\cite{flux2}, FireRed-Image-Edit-1.1~\cite{firered-1.0}, and JoyAI-Image-Edit~\cite{song2026joyai}, both with and without CRIE-Agent. We also compare with three editing agents, ReasonEdit-S~\cite{yin2026reasonedit}, IntentEdit~\cite{zhang2026intentedit} and RePlan~\cite{RePlan}, using their official workflows. To study how instruction processing affects editing, we compare punctuation splitting, VLM splitting, VLM reorganization, and the Intent Agent.

\paragraph{Evaluation metrics.}
We use both Qwen3-VL-32B-Instruct~\cite{qwen3vl2025} and GLM-4.6V-Flash~\cite{hong2025glm} to evaluate object correctness, attribute accuracy and spatial relations. The metrics assess whether the edited image satisfies the annotated final intent, with relative size included in attribute accuracy. Complete success is the percentage of test samples for which every required object, attribute, and spatial relation check receives a valid answer and is satisfied. The main text reports results from Qwen3-VL-32B-Instruct, while the GLM-4.6V-Flash results are provided in Appendix~\ref{sec:independent_judge}.

\subsection{Results and Analysis}
\label{sec:results_analysis}
CRIE-Agent improves complete editing success across all five base editors, from 0\% with direct editing to 27.5--50.0\% (Table~\ref{tab:main}). With Qwen-Image-2.1 and FLUX.2-klein-4B, it reaches 50.0\% and 27.5\%, respectively. Object, attribute, and relation accuracy improve with every editor, showing that the framework helps realize the final requirements across different editing backends. Existing editing agents achieve 0--1.0\% complete success under their evaluated workflows. Figure~\ref{fig:qualitative} illustrates the corresponding improvements in object attributes and spatial relations. Additional completed examples are provided in Appendix Figure~\ref{fig:additional_qualitative}. Evaluation with GLM-4.6V-Flash also supports CRIE-Agent's improvements across editors. Detailed results are provided in the appendix.

\begin{table*}[t]
\centering
\caption{Execution configurations on the CRIE-S test set with Qwen-Image-Edit 2511. CRIE-Agent uses reference-first scheduling with $K=2$. Bold values indicate the highest score in each column.}
\label{tab:abl_exec}
\begingroup
\small
\renewcommand{\arraystretch}{1.08}
\begin{tabular}{@{}lcccccccc@{}}
\toprule
& \multicolumn{3}{c}{Execution Agent} & \multicolumn{4}{c}{Editing quality} \\
\cmidrule(lr){2-4}\cmidrule(lr){5-8}
Method & Verification & Layout & Block Grouping & Object $\uparrow$ & Attribute $\uparrow$ & Relations $\uparrow$  & Success $\uparrow$ \\
\midrule
Intent Agent & -- & -- & -- & 92.0 & 88.0 & 69.4 & 11.0 \\
\quad + Verification & $\checkmark$ & -- & -- & 94.1 & 90.5 & 75.9 & 21.5 \\
\quad + Layout & $\checkmark$ & $\checkmark$ & -- & 92.9 & 89.7 & 81.5 & 26.5 \\
\midrule
\rowcolor{crieblue}
\textbf{CRIE-Agent (Ours)} & $\checkmark$ & $\checkmark$ & $\checkmark$ & \textbf{94.6} & \textbf{92.0} & \textbf{87.5} & \textbf{39.5} \\
\bottomrule
\end{tabular}
\endgroup
\end{table*}

\begin{table*}[t]
\centering
\caption{Comparison of block scheduling with $K=2$ on CRIE-S. Scores are percentages, reported as mean $\pm$ sample standard deviation over three seeds. Bold values indicate the highest mean in each column.}
\label{tab:traversal}
\begingroup
\small
\renewcommand{\arraystretch}{1.08}
\begin{tabular}{@{}lcccccc@{}}
\toprule
Strategy & $K$ & Object $\uparrow$ & Attribute $\uparrow$ & Relations $\uparrow$  & Success $\uparrow$ \\
\midrule
Random & 2 & $93.1\pm0.5$ & $90.4\pm0.7$ & $79.5\pm1.3$ & $23.2\pm1.4$ \\
Left-to-right & 2 & $94.4\pm0.4$ & $91.6\pm0.4$ & $81.9\pm1.2$ & $30.8\pm1.2$ \\
Right-to-left & 2 & $93.4\pm0.5$ & $90.3\pm0.3$ & $83.6\pm1.6$ & $27.5\pm3.5$ \\
\rowcolor{crieblue}
\textbf{Reference-first (Ours)} & 2 & $\mathbf{94.9\pm0.6}$ & $\mathbf{92.2\pm0.2}$ & $\mathbf{87.3\pm0.4}$  & $\mathbf{39.3\pm4.3}$ \\
\bottomrule
\end{tabular}
\endgroup
\end{table*}

\subsection{Ablation Studies}

\paragraph{Intent Agent.}
Learned intent recovery improves how the editor follows the final requirements (Table~\ref{tab:abl_intent}). Splitting an instruction into shorter commands preserves its revision history, whereas VLM reorganization attempts to remove canceled or superseded edits. Single-step execution of the reorganized requirements performs better than sequential execution. Under the same single-step setting, the Intent Agent doubles complete success from 5.5\% to 11.0\%. Its recovered intents are semantically consistent with the annotations on all test instructions, showing that it supplies the editor with the correct final requirements on this test set.

We further evaluate generalization to examine whether the Intent Agent learns to recover the final intent beyond the fixed instruction patterns in CRIE-S. These experiments vary instruction phrasing, object counts, and revision history length. The results show that the Intent Agent remains effective under these changes, improving single-step editing success by 9.3 percentage points over VLM reorganization on naturally rephrased instructions (Appendix~\ref{sec:intent_generalization}). Additional comparisons with 8B and 32B schema-prompted models under extended output budgets are provided in Appendix~\ref{sec:stronger_intent_baselines}.

\paragraph{Is Intent Recovery Sufficient for Editing?}
We investigate whether recovering the final intent is sufficient for complex image editing, or whether an explicit execution strategy is still needed. Given the same intent recovered by the Intent Agent, we compare directly editing the image in one step, splitting the requirements into fine-grained sequential edits, and using our Execution Agent to coordinate their execution (Appendix Table~\ref{tab:recovered_intent}). Across three editors, CRIE-Agent improves complete success by 19.0--23.0 percentage points over fine-grained editing, while using fewer editor calls per sample (6.88--7.46 versus 10.51). These results show that recovering the intent or simply increasing editing steps is insufficient to match our coordinated execution strategy.

\paragraph{Execution Agent.}
The Execution Agent improves visual realization of the recovered requirements (Table~\ref{tab:abl_exec}). Verification checks complete candidates and guides retries, Layout supplies region hints, and Block Grouping generates referenced objects before their dependents in smaller editing steps. Together, these components raise complete success from 11.0\% to 39.5\% and improve relation accuracy to 87.5\%. 

Figure~\ref{fig:execution_qualitative} illustrates the complexity of visual execution after intent recovery. The editor must jointly realize multiple objects, their attributes, and interdependent spatial relations. CRIE-Agent addresses this challenge by checking completed candidates and organizing edits around reference objects, with layout hints providing additional spatial guidance when needed. The full framework satisfies the highlighted constraints and avoids the extra or deformed objects shown in the comparison outputs.

\paragraph{Block scheduling.}
We first compare different block scheduling strategies. As shown in Table~\ref{tab:traversal}, our reference-first scheduling outperforms random and fixed-direction scheduling. By generating referenced objects before their dependents, it establishes spatial anchors for subsequent edits, helping the editor satisfy inter-object relations. We also compare block capacities $K\in\{1,2,3,4\}$ and select $K=2$ as the default configuration; detailed results are provided in Appendix~\ref{sec:block_capacity_results}.

\subsection{Limitations}
VLM verification can miss visible objects or reject satisfied spatial relations (Appendix~\ref{sec:additional_cases}), leading to unnecessary retries or affecting candidate selection. A direction for further improvement is to enable more accurate, fine-grained inspection of edited images, providing richer feedback for execution and selection.
\section{Conclusion}
\label{Conclusion}

We introduced complex redundant-instruction image editing, which requires resolving revisions within a single request and jointly realizing the resulting object attributes and relations. CRIE-S pairs editing histories with exact final-intent annotations to study this task. CRIE-Agent separates final-intent recovery from visual execution, using the recovered requirements to guide reference-first editing and verification. Across five image-editing backends, it raises complete editing success from 0\% to 27.5--50.0\%. These results establish the value of resolving what should remain before organizing how the edits should be executed.
\section{LLM Usage Statement}
\label{sec:llm_usage_statement}

We used large language models (LLMs) as assistants throughout this work, including for the initial literature survey and reference search, writing and debugging code, and drafting, polishing, and proofreading the manuscript. All LLM-generated content, including code, cited references, and text, was reviewed and verified by the authors, who take full responsibility for the content of this paper.
{
    \small
    \bibliographystyle{ieeenat_fullname}
    \bibliography{main}

@String(CVPR= {IEEE Conf. Comput. Vis. Pattern Recog.})

@String(ICCV= {Int. Conf. Comput. Vis.})

@String(ECCV= {Eur. Conf. Comput. Vis.})

@String(ACMMM= {ACM Int. Conf. Multimedia})

@String(ICLR = {Int. Conf. Learn. Represent.})

@String(AAAI = {AAAI})

@String(CVPR  = {CVPR})

@String(ICCV  = {ICCV})

@String(ECCV  = {ECCV})

@String(ACMMM = {ACM MM})

@String(ICLR  = {ICLR})

@inproceedings{instructpix2pix,
  author       = {Tim Brooks and
                  Aleksander Holynski and
                  Alexei A. Efros},
  title        = {InstructPix2Pix: Learning to Follow Image Editing Instructions},
  booktitle    = {{CVPR}},
  pages        = {18392--18402},
  year         = {2023},
}

@inproceedings{emuedit,
  author       = {Shelly Sheynin and
                  Adam Polyak and
                  Uriel Singer and
                  Yuval Kirstain and
                  Amit Zohar and
                  Oron Ashual and
                  Devi Parikh and
                  Yaniv Taigman},
  title        = {Emu Edit: Precise Image Editing via Recognition and Generation Tasks},
  booktitle    = {{CVPR}},
  pages        = {8871--8879},
  year         = {2024},
}

@inproceedings{anyedit,
  author       = {Qifan Yu and
                  Wei Chow and
                  Zhongqi Yue and
                  Kaihang Pan and
                  Yang Wu and
                  Xiaoyang Wan and
                  Juncheng Li and
                  Siliang Tang and
                  Hanwang Zhang and
                  Yueting Zhuang},
  title        = {AnyEdit: Mastering Unified High-Quality Image Editing for Any Idea},
  booktitle    = {{CVPR}},
  pages        = {26125--26135},
  year         = {2025},
}

@inproceedings{mgie,
  author       = {Tsu{-}Jui Fu and
                  Wenze Hu and
                  Xianzhi Du and
                  William Yang Wang and
                  Yinfei Yang and
                  Zhe Gan},
  title        = {Guiding Instruction-based Image Editing via Multimodal Large Language
                  Models},
  booktitle    = {{ICLR}},
  year         = {2024},
  pages        = {1--24},
}

@inproceedings{xplanner,
  author       = {Chun{-}Hsiao Yeh and
                  Yilin Wang and
                  Nanxuan Zhao and
                  Richard Zhang and
                  Yuheng Li and
                  Yi Ma and
                  Krishna Kumar Singh},
  title        = {Beyond Simple Edits: X-Planner for Complex Instruction-Based Image
                  Editing},
  booktitle    = {{AAAI}},
  pages        = {11991--11999},
  year         = {2026},
}

@article{flux1,
  title={Flux. 1 kontext: Flow matching for in-context image generation and editing in latent space},
  author={Labs, Black Forest and Batifol, Stephen and Blattmann, Andreas and Boesel, Frederic and Consul, Saksham and Diagne, Cyril and Dockhorn, Tim and English, Jack and English, Zion and Esser, Patrick and others},
  journal={arXiv preprint arXiv:2506.15742},
  pages        = {1--19},
  year={2025}
}

@article{qwenimage,
  title={Qwen-image technical report},
  author={Wu, Chenfei and Li, Jiahao and Zhou, Jingren and Lin, Junyang and Gao, Kaiyuan and Yan, Kun and Yin, Sheng-ming and Bai, Shuai and Xu, Xiao and Chen, Yilei and others},
  journal={arXiv preprint arXiv:2508.02324},
  pages        = {1--46},
  year={2025}
}

@inproceedings{SmartEdit,
  author       = {Yuzhou Huang and
                  Liangbin Xie and
                  Xintao Wang and
                  Ziyang Yuan and
                  Xiaodong Cun and
                  Yixiao Ge and
                  Jiantao Zhou and
                  Chao Dong and
                  Rui Huang and
                  Ruimao Zhang and
                  others},
  title        = {SmartEdit: Exploring Complex Instruction-Based Image Editing with Multimodal Large Language Models},
  booktitle    = {CVPR},
  year         = {2024},
  pages        = {8362–8371},
}

@inproceedings{GenArtist,
  author       = {Zhenyu Wang and
                  Aoxue Li and
                  Zhenguo Li and
                  Xihui Liu},
  title        = {GenArtist: Multimodal {LLM} as an Agent for Unified Image Generation
                  and Editing},
  booktitle    = {NeurIPS},
  year         = {2024},
  pages        = {128374–128395},
}

@inproceedings{RePlan,
  author       = {Tianyuan Qu and
                  Lei Ke and
                  Xiaohang Zhan and
                  Longxiang Tang and
                  Yuqi Liu and
                  Bohao Peng and
                  Bei Yu and
                  Dong Yu and
                  Jiaya Jia},
  title        = {RePlan: Reasoning-Guided Region Planning for Complex Instruction-Based
                  Image Editing},
  booktitle    = {{ECCV}},
  pages        = {616--631},
  year         = {2026},
}

@misc{flux2,
  author = {{Black Forest Labs}},
  title = {{FLUX.2 [klein]}: Towards Interactive Visual Intelligence},
  year = {2026},
  url = {https://bfl.ai/blog/flux2-klein-towards-interactive-visual-intelligence},
  note = {Accessed: 2026-01-15}
}

@misc{qwen2026image21,
  author = {{Qwen Team}},
  title = {{Qwen-Image-2.1}},
  year = {2026},
  url = {https://github.com/QwenLM/Qwen-Image-2.1},
  note = {Accessed: 2026-09-30}
}

@inproceedings{yao2023react,
  author = {Shunyu Yao and Jeffrey Zhao and Dian Yu and Nan Du and Izhak Shafran and Karthik Narasimhan and Yuan Cao},
  title = {{ReAct}: Synergizing Reasoning and Acting in Language Models},
  booktitle = {ICLR},
  year = {2023},
  pages = {1--33}
}

@inproceedings{shen2023hugginggpt,
  author = {Yongliang Shen and Kaitao Song and Xu Tan and Dongsheng Li and Weiming Lu and Yueting Zhuang},
  title = {{HuggingGPT}: Solving {AI} Tasks with {ChatGPT} and its Friends in {Hugging Face}},
  booktitle = {NeurIPS},
  pages = {38154–38180},
  year = {2023},
}

@inproceedings{agashe2025agents,
  author = {Saaket Agashe and Jiuzhou Han and Shuyu Gan and Jiachen Yang and Ang Li and Xin Eric Wang},
  title = {{Agent S}: An Open Agentic Framework that Uses Computers Like a Human},
  booktitle = {ICLR},
  year = {2025},
  pages = {1--23},
}

@inproceedings{wang2025openhands,
  author       = {Xingyao Wang and
                  Boxuan Li and
                  Yufan Song and
                  Frank F. Xu and
                  Xiangru Tang and
                  Mingchen Zhuge and
                  Jiayi Pan and
                  Yueqi Song and
                  Bowen Li and
                  Jaskirat Singh and
                  others},
  title = {{OpenHands}: An Open Platform for {AI} Software Developers as Generalist Agents},
  booktitle = {ICLR},
  year = {2025},
  pages = {1--38}
}

@inproceedings{li2024styletokenizer,
  author = {Wen Li and Muyuan Fang and Cheng Zou and Biao Gong and Ruobing Zheng and Meng Wang and Jingdong Chen and Ming Yang},
  title = {{StyleTokenizer}: Defining Image Style by a Single Instance for Controlling Diffusion Models},
  booktitle = {ECCV},
  pages = {110--126},
  year = {2024},
}

@inproceedings{lora,
  author       = {Edward J. Hu and
                  Yelong Shen and
                  Phillip Wallis and
                  Zeyuan Allen{-}Zhu and
                  Yuanzhi Li and
                  Shean Wang and
                  Lu Wang and
                  Weizhu Chen},
  title        = {LoRA: Low-Rank Adaptation of Large Language Models},
  booktitle    = {ICLR},
  year         = {2022},
  pages        = {1--26},
}

@article{firered-1.0,
  title={FireRed-Image-Edit-1.0 Technical Report}, 
  author={Super Intelligence Team},
  journal={arXiv preprint arXiv:2602.13344},
  year={2026},
  pages={1--34}
}

@article{song2026joyai,
  title={JoyAI-Image: Awaking Spatial Intelligence in Unified Multimodal Understanding and Generation},
  author={Song, Lin and Li, Wenbo and Ma, Guoqing and Tang, Wei and Wang, Bo and Zhang, Yuan and Yang, Yijun and Xiao, Yicheng and Liu, Jianhui and Zhang, Yanbing and others},
  journal={arXiv preprint arXiv:2605.04128},
  year={2026},
  pages={1--48}
}

@inproceedings{yin2026reasonedit,
  author       = {Fukun Yin and
                  Shiyu Liu and
                  Yucheng Han and
                  Zhibo Wang and
                  Peng Xing and
                  Rui Wang and
                  Wei Cheng and
                  Yingming Wang and
                  Aojie Li and
                  Zixin Yin and
                  others},
  title        = {ReasonEdit: Towards Reasoning-Enhanced Image Editing Models},
  booktitle    = {CVPR},
  pages        = {23283--23293},
  year         = {2026},
}

@inproceedings{zhang2026intentedit,
  title={IntentEdit: Multi-Agent Reasoning for Intent-Driven Complex Image Editing},
  author={Zhang, Yuxuan and Huang, Shijia and Wang, Liwei},
  booktitle={CVPR Findings},
  pages={8776--8785},
  year={2026}
}

@article{qwen3vl2025,
  title={Qwen3-VL Technical Report},
  author={Bai, Shuai and Cai, Yuxuan and Chen, Ruizhe and Chen, Keqin and Chen, Xionghui and Cheng, Zesen and Deng, Lianghao and Ding, Wei and Gao, Chang and Ge, Chunjiang and others},
  journal={arXiv preprint arXiv:2511.21631},
  year={2025},
  pages={1--42}
}

@article{qwen25vl2025,
  title={Qwen2.5-VL Technical Report},
  author       = {Shuai Bai and
                  Keqin Chen and
                  Xuejing Liu and
                  Jialin Wang and
                  Wenbin Ge and
                  Sibo Song and
                  Kai Dang and
                  Peng Wang and
                  Shijie Wang and
                  Jun Tang and
                  others},
  journal={arXiv preprint arXiv:2502.13923},
  year={2025},
  pages={1--23}
}

@article{zhu2025internvl3,
  title={Internvl3: Exploring advanced training and test-time recipes for open-source multimodal models},
  author={Zhu, Jinguo and Wang, Weiyun and Chen, Zhe and Liu, Zhaoyang and Ye, Shenglong and Gu, Lixin and Tian, Hao and Duan, Yuchen and Su, Weijie and Shao, Jie and others},
  journal={arXiv preprint arXiv:2504.10479},
  year={2025},
  pages={1--27}
}

@article{deng2025bagel,
  title={Emerging properties in unified multimodal pretraining},
  author={Deng, Chaorui and Zhu, Deyao and Li, Kunchang and Gou, Chenhui and Li, Feng and Wang, Zeyu and Zhong, Shu and Yu, Weihao and Nie, Xiaonan and Song, Ziang and others},
  journal={arXiv preprint arXiv:2505.14683},
  year={2025},
  pages={1--37}
}

@inproceedings{Zhang2023MagicBrush,
  author       = {Kai Zhang and
                  Lingbo Mo and
                  Wenhu Chen and
                  Huan Sun and
                  Yu Su},
  title        = {MagicBrush: {A} Manually Annotated Dataset for Instruction-Guided
                  Image Editing},
  booktitle    = {NeurIPS},
  year         = {2023},
  pages        = {31428--31449}
}

@inproceedings{Hui2025HQEdit,
  author       = {Mude Hui and
                  Siwei Yang and
                  Bingchen Zhao and
                  Yichun Shi and
                  Heng Wang and
                  Peng Wang and
                  Cihang Xie and
                  Yuyin Zhou},
  title        = {HQ-Edit: {A} High-Quality Dataset for Instruction-based Image Editing},
  booktitle    = {ICLR},
  year         = {2025},
  pages        = {1--22}
}

@inproceedings{Ma2024I2EBench,
  author       = {Yiwei Ma and
                  Jiayi Ji and
                  Ke Ye and
                  Weihuang Lin and
                  Zhibin Wang and
                  Yonghan Zheng and
                  Qiang Zhou and
                  Xiaoshuai Sun and
                  Rongrong Ji},
  title        = {I2EBench: {A} Comprehensive Benchmark for Instruction-based Image
                  Editing},
  booktitle    = {NeurIPS},
  year         = {2024},
  pages        = {41494--41516}
}

@inproceedings{Ye2025ImgEdit,
  author       = {Yang Ye and
                  Xianyi He and
                  Zongjian Li and
                  Bin Lin and
                  Shenghai Yuan and
                  Zhiyuan Yan and
                  Bohan Hou and
                  Li Yuan},
  title        = {ImgEdit: {A} Unified Image Editing Dataset and Benchmark},
  booktitle    = {NeurIPS},
  year         = {2025},
  pages        = {1--31}
}

@inproceedings{Wang2025ComplexBenchEdit,
  author       = {Chenglin Wang and
                  Yucheng Zhou and
                  Qianning Wang and
                  Zhe Wang and
                  Kai Zhang},
  title        = {ComplexBench-Edit: Benchmarking Complex Instruction-Driven Image Editing
                  via Compositional Dependencies},
  booktitle    = {ACMMM},
  pages        = {13391--13397},
  year         = {2025},
}

@article{Yang2026ComplexEdit,
  author       = {Siwei Yang and
                  Mude Hui and
                  Bingchen Zhao and
                  Yuyin Zhou and
                  Nataniel Ruiz and
                  Cihang Xie},
  title        = {Complex-Edit: CoT-Like Instruction Generation
                  for Complexity-Controllable Image Editing Benchmark},
  journal      = {Trans. Mach. Learn. Res.},
  volume       = {2026},
  year         = {2026},
  pages        = {1--25}
}

@inproceedings{Jia2026CompBench,
  author       = {Bohan Jia and
                  Wenxuan Huang and
                  Yuntian Tang and
                  Junbo Qiao and
                  Jincheng Liao and
                  Shaosheng Cao and
                  Fei Zhao and
                  Feng Zhao and
                  Zhouhong Gu and
                  Zhenfei Yin and
                  others},
  title        = {CompBench: Benchmarking Complex Instruction-guided Image Editing},
  booktitle    = {CVPR},
  pages        = {1112--1122},
  year         = {2026},
}

@article{Basu2023EditVal,
  title={Editval: Benchmarking diffusion based text-guided image editing methods},
  author={Basu, Samyadeep and Saberi, Mehrdad and Bhardwaj, Shweta and Chegini, Atoosa Malemir and Massiceti, Daniela and Sanjabi, Maziar and Hu, Shell Xu and Feizi, Soheil},
  journal={arXiv preprint arXiv:2310.02426},
  year={2023}
}

@inproceedings{bai2026uniediti,
  author       = {Chengyu Bai and
                  Jintao Chen and
                  Xiang Bai and
                  Yilong Chen and
                  Qi She and
                  Ming Lu and
                  Shanghang Zhang},
  title        = {UniEdit-I: Training-free Image Editing for Unified {VLM} via Iterative
                  Understanding, Editing and Verifying},
  booktitle    = {CVPR},
  pages        = {29750--29759},
  year         = {2026},
}

@inproceedings{singh2023divide,
  author       = {Jaskirat Singh and
                  Liang Zheng},
  title        = {Divide, Evaluate, and Refine: Evaluating and Improving Text-to-Image
                  Alignment with Iterative {VQA} Feedback},
  booktitle    = {NeurIPS},
  year         = {2023},
  pages        = {70799--70811}
}

@inproceedings{hu2023tifa,
  author       = {Yushi Hu and
                  Benlin Liu and
                  Jungo Kasai and
                  Yizhong Wang and
                  Mari Ostendorf and
                  Ranjay Krishna and
                  Noah A. Smith},
  title        = {{TIFA:} Accurate and Interpretable Text-to-Image Faithfulness Evaluation
                  with Question Answering},
  booktitle    = {ICCV},
  pages        = {20349--20360},
  year         = {2023},
}

@inproceedings{cho2024dsg,
  author       = {Jaemin Cho and
                  Yushi Hu and
                  Jason M. Baldridge and
                  Roopal Garg and
                  Peter Anderson and
                  Ranjay Krishna and
                  Mohit Bansal and
                  Jordi Pont{-}Tuset and
                  Su Wang},
  title        = {Davidsonian Scene Graph: Improving Reliability in Fine-grained Evaluation
                  for Text-to-Image Generation},
  booktitle    = {ICLR},
  year         = {2024},
  pages        = {1--21}
}

@inproceedings{ku2024viescore,
  author       = {Max Ku and
                  Dongfu Jiang and
                  Cong Wei and
                  Xiang Yue and
                  Wenhu Chen},
  title        = {VIEScore: Towards Explainable Metrics for Conditional Image Synthesis
                  Evaluation},
  booktitle    = {ACL},
  pages        = {12268--12290},
  year         = {2024},
}

@inproceedings{saxon2024ts2,
  author       = {Michael Saxon and
                  Fatima Jahara and
                  Mahsa Khoshnoodi and
                  Yujie Lu and
                  Aditya Sharma and
                  William Yang Wang},
  title        = {Who Evaluates the Evaluations? Objectively Scoring Text-to-Image Prompt
                  Coherence Metrics with T2IScoreScore {(TS2)}},
  booktitle    = {NeurIPS},
  year         = {2024},
  pages        = {85630--85657}
}

@inproceedings{zeng2026robustsequential,
  author       = {Zilai Zeng and
                  Mingdeng Cao and
                  Zijie Li and
                  Xiaochen Lian and
                  Yichun Shi and
                  Peihao Zhu and
                  Chen Sun and
                  Peng Wang},
  title        = {Towards Robust Sequential Decomposition for Complex Image Editing},
  booktitle    = {CVPR},
  pages        = {38101--38110},
  year         = {2026},
}

@article{zhao2026adaptivereformulation,
  title={Making image editing easier via adaptive task reformulation with agentic executions},
  author={Zhao, Bo and Guo, Kairui and Du, Runnan and Sun, Haiyang and Wang, Pengshan and Yang, Huan and Gai, Kun and Cao, Yixin and Ji, Wei},
  journal={arXiv preprint arXiv:2604.15917},
  year={2026}
}

@article{hong2025glm,
  title={Glm-4.5 v and glm-4.1 v-thinking: Towards versatile multimodal reasoning with scalable reinforcement learning},
  author={Hong, Wenyi and Yu, Wenmeng and Gu, Xiaotao and Wang, Guo and Gan, Guobing and Tang, Haomiao and Cheng, Jiale and Qi, Ji and Ji, Junhui and Pan, Lihang and others},
  journal={arXiv preprint arXiv:2507.01006},
  year={2025}
}
}

\clearpage
\setcounter{page}{1}
\maketitlesupplementary
\appendix

\begin{figure*}[t]
\centering
\includegraphics[width=0.9\textwidth]{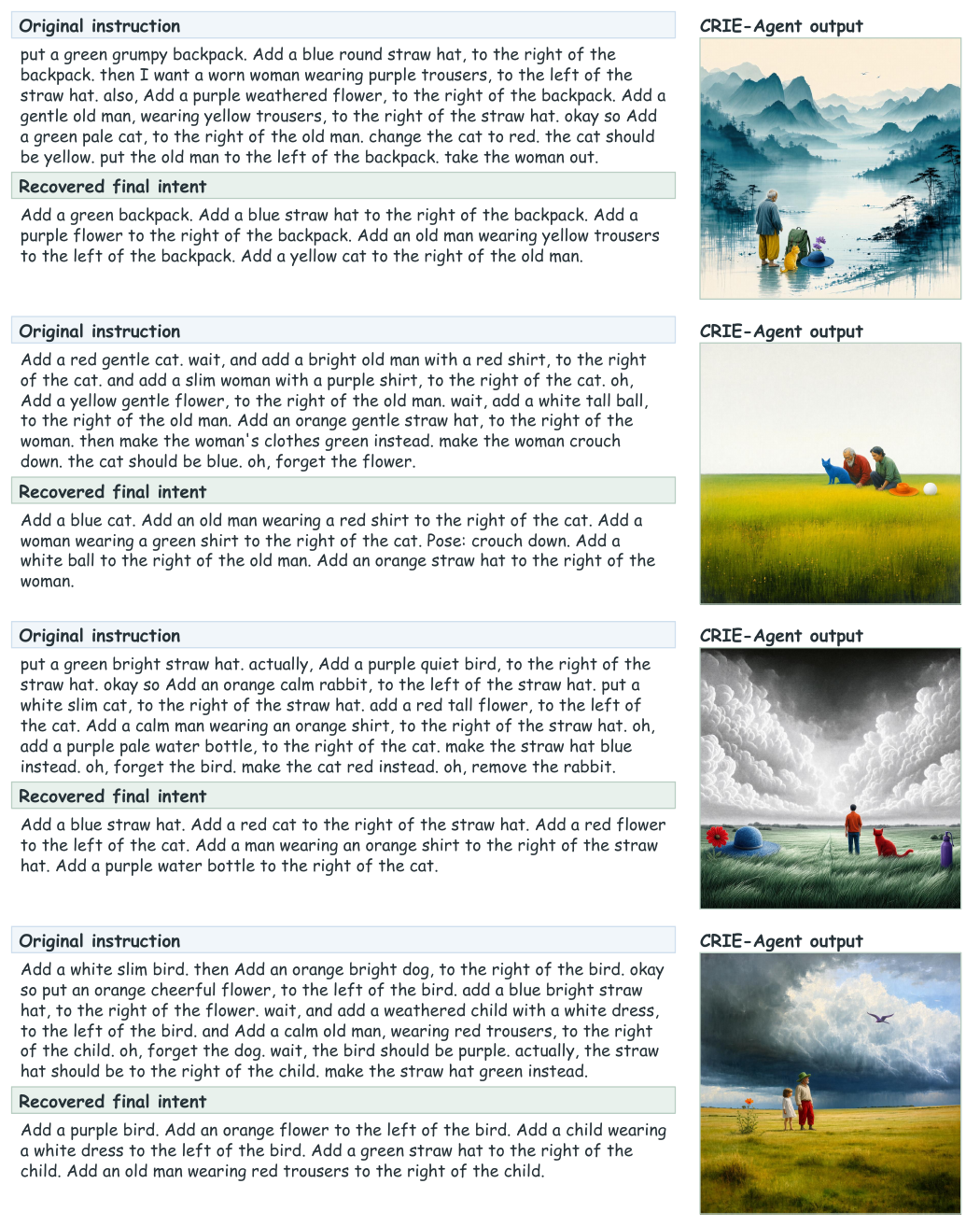}
\caption{Additional qualitative examples of CRIE-Agent. }
\label{fig:additional_qualitative}
\end{figure*}

\section{Theoretical Analysis}
\label{sec:theory}

We analyze Block Grouping in the Execution Agent of CRIE-Agent. This component turns the object records predicted by the Intent Agent into an ordered plan of editor calls. 
We establish that Block Grouping preserves reference dependencies and bound the number of editor calls required to execute one complete plan.

\paragraph{Problem setup.}
Fix the predicted records $\hat Z$ and their reference graph
$G_{\mathrm{ref}}=(V,E_{\mathrm{ref}})$, as defined in
Section~\ref{sec:block_grouping}. Assume that the graph is a
nonempty DAG and the block capacity $K\geq1$ is an integer.
Let $N=|V|$, and let $H$ be the number of vertices on a longest
directed path. Block Grouping computes the reference depths
$d(v)$ and layers $L_h$, then divides each layer into full
$K$-object blocks and at most one smaller block. The resulting
plan $\Pi=(B_1,\ldots,B_T)$ requires $T$ editor calls in one
complete attempt, with one call per block.

\paragraph{Comparison criterion.}
A feasible plan partitions the same objects into nonempty
blocks of at most $K$ objects and places each reference in
an earlier block than its dependents. Formally, its block
indices satisfy $b(u)<b(v)$ for every
$(u,v)\in E_{\mathrm{ref}}$, where $b(v)$ denotes the block
containing $v$. Let $T^*$ be the minimum number of editor
calls among all such plans, including those that combine
objects from different layers.

\begin{crietheorem}[Dependency preservation]
\label{thm:reference_feasibility}
In the plan $\Pi$ returned by Block Grouping, every predicted object record belongs to exactly one nonempty block of at most $K$ objects, and each reference edge satisfies $b(u)<b(v)$. The depth $d(v)$ equals the maximum number of vertices on a directed path ending at $v$, and the construction has exactly $H$ layers.
\end{crietheorem}

The reference graph retains which record provides the reference for each dependent description. Ordering its layers places the corresponding editing operation in an earlier block. Thus, the plan places every reference object before the objects that depend on it.

\begin{crietheorem}[Editor-call budget of Block Grouping]
\label{thm:call_efficiency}
Let $T$ count the calls to $\mathcal F$ in one execution of $\Pi$ under Eq.~\eqref{eq:execute}. Write $n_h=|L_h|$ and $U=H+\lfloor(N-H)/K\rfloor$. Then
\begin{equation}
\begin{aligned}
T&=\sum_{h=1}^{H}\left\lceil\frac{n_h}{K}\right\rceil\leq U,\\
\frac{T}{T^*}
&\leq\frac{U}{\max\{H,\lceil N/K\rceil\}}\leq2-\frac1K.
\end{aligned}
\label{eq:layer_call_count}
\end{equation}
For five actual scheduled records and $K=2$, the sharper bound is $T\leq(4/3)T^*$, and the factor $4/3$ is tight.
\end{crietheorem}

Block packing controls the calls within each dependency layer, while reference depth gives a lower bound on successive calls. For a fixed graph, the call count is nonincreasing as $K$ increases. When $K\geq\max_h n_h$, packing reaches the depth limit $T=T^*=H$. 
The bound characterizes the editor calls within one complete attempt; the capacity study evaluates the editing quality obtained with different values of $K$.
The bounds concern one complete attempt and exclude verification and retry costs.

\subsection{Proofs for Block Grouping}
\label{sec:planning_properties}

The proofs analyze the depth layers and block packing used by CRIE-Agent. They establish dependency preservation and bounds on the calls made to the editor, including the five-object case and capacity boundaries.

\paragraph{Proof of Theorem~\ref{thm:reference_feasibility}.}
Block Grouping applies the depth recurrence and layer packing of Section~\ref{sec:block_grouping} to the reference graph of the fixed predicted records $\hat Z$.

Under the depth recurrence, a source has depth $1$. For any other vertex, the depth is $1$ plus the maximum predecessor depth. Induction in a topological order shows that this depth is the maximum number of vertices on a directed path ending there. Every path into a nonsource ends through a predecessor, and appending the vertex adds one. Conversely, a predecessor attaining the maximum supplies a path attaining this depth. The largest depth is therefore $H$.

A depth-$1$ vertex is removed at the first zero-indegree removal step. For a vertex of depth $h>1$, all predecessors have depth at most $h-1$, and at least one has depth $h-1$. Induction over the removal steps shows that it is removed exactly at step $h$. Thus depth equals layer index, and the removal procedure has $H$ nonempty layers. Every edge $(u,v)$ satisfies $d(v)\geq d(u)+1$, so its endpoints are in strictly ordered layers.

Each vertex has exactly one depth. The layers therefore partition the predicted records, and the consecutive full blocks and optional final smaller blocks of Block Grouping partition each layer without duplication. The blocks are nonempty and have size at most $K$. Completing every layer before the next places every reference in a strictly earlier block than its dependent, giving $b(u)<b(v)$. \hfill$\square$

\paragraph{Proof of Theorem~\ref{thm:call_efficiency}.}
Eq.~\eqref{eq:execute} applies $\mathcal F$ once per block $t=1,\ldots,T$ in the plan returned by Block Grouping. Theorem~\ref{thm:reference_feasibility} establishes feasibility. Packing the $n_h$ records in each layer into full $K$-object blocks and at most one smaller block gives exactly $\lceil n_h/K\rceil$ calls. Summing over the layers yields Eq.~\eqref{eq:layer_call_count}.

For every positive integer $n$, division by integer $K\geq1$ gives
\begin{equation}
\left\lceil\frac nK\right\rceil
=1+\left\lfloor\frac{n-1}{K}\right\rfloor.
\end{equation}
The sum of floors is an integer no larger than the sum of their underlying quantities and is therefore at most the floor of that sum. Since $\sum_h n_h=N$,
\begin{equation}
\begin{aligned}
T=H+\sum_{h=1}^{H}\left\lfloor\frac{n_h-1}{K}\right\rfloor \leq H+\left\lfloor\frac{N-H}{K}\right\rfloor=U.
\end{aligned}
\label{eq:planning_integer_bound}
\end{equation}
Every feasible comparison plan has at least $H$ blocks by strict precedence along a longest dependency chain and at least $\lceil N/K\rceil$ blocks by capacity. Hence, with $L=\max\{H,\lceil N/K\rceil\}>0$, we have $T^*\geq L$ and $T/T^*\leq U/L$. Finally,
\begin{equation}
U\leq\frac NK+\left(1-\frac1K\right)H.
\end{equation}
Using $L\geq N/K$ and $L\geq H$ proves the entire ratio chain:
\begin{equation}
\begin{aligned}
\frac{T}{T^*}&\leq\frac UL \leq\frac{N}{KL}+\left(1-\frac1K\right)\frac HL \leq2-\frac1K.
\end{aligned}
\label{eq:planning_ratio_bound}
\end{equation}

\paragraph{Five-object specialization and tightness.}
For $N=5$ and $K=2$, the same argument gives
\[
\frac{T}{T^*}
\leq\frac{H+\lfloor(5-H)/2\rfloor}{\max\{H,3\}}.
\]
For $H=1,2,3,4,5$, the right-hand side equals, respectively, $1,1,4/3,1,1$. Thus $T\leq(4/3)T^*$ for the stated five-object case.

For tightness, consider five predicted records with only the edges $a\to b$ and $b\to c$, and isolated records $x,y$. Block Grouping yields layers $\{a,x,y\}$, $\{b\}$, and $\{c\}$. For an admissible within-layer order, compare its plan with a feasible cross-layer plan:
\begin{equation}
\begin{aligned}
\Pi_{\mathrm{layer}}&=(\{a,x\},\{y\},\{b\},\{c\}),\\
\Pi_{\mathrm{compare}}&=(\{a,x\},\{b,y\},\{c\}).
\end{aligned}
\label{eq:planning_tight_example}
\end{equation}
The count for Block Grouping is four regardless of the within-layer order. 
The comparison plan satisfies the same capacity and strict dependencies in three calls. 
Every feasible plan has at least three calls because of the three-vertex chain, so $T^*=3$ and $T/T^*=4/3$. This completes the proof of Theorem~\ref{thm:call_efficiency}. \hfill$\square$

\paragraph{Capacity and boundary cases.}
For a fixed reference graph, each term $\lceil n_h/K\rceil$ is nonincreasing in integer $K$. If $K\geq\max_h n_h$, each layer fits in one block, so $T=H$. The longest-path lower bound then gives $T^*=H$. In particular, this holds for $K\geq N$. For $K=1$, every feasible plan has $N$ blocks and the construction uses exactly $N$, so $T=T^*=N$.

\section{Related Work}
\label{sec:related_work}

\paragraph{Image Editing Data and Benchmarks.}
MagicBrush and HQ-Edit provide annotated and synthesized instruction-guided editing data~\cite{Zhang2023MagicBrush,Hui2025HQEdit}. Benchmarks such as EditVal, I2EBench, and CompBench benchmark diverse editing operations and increasingly complex instructions~\cite{Basu2023EditVal,Ma2024I2EBench,Jia2026CompBench}. ImgEdit includes multi-turn editing and version backtracking~\cite{Ye2025ImgEdit}, while ComplexBench-Edit studies compositional dependencies and Complex-Edit varies instruction complexity~\cite{Wang2025ComplexBenchEdit,Yang2026ComplexEdit}. CRIE-S focuses on revisions within a single request, pairing the instruction with its atomic editing history and exact final object-level intent. This setting evaluates both which requirements remain valid and how well their attributes and relations are realized in the image.

\paragraph{Planning and Agents for Image Editing.}
Multimodal language models can interpret editing requests and provide guidance to image editors~\cite{mgie,SmartEdit}. Agent-based methods further organize complex edits through planning and tool use~\cite{GenArtist,xplanner,RePlan,zhang2026intentedit}. Recent work explores reasoning-enhanced editing~\cite{yin2026reasonedit}, learned sequential decomposition~\cite{zeng2026robustsequential}, and adaptive task reformulation~\cite{zhao2026adaptivereformulation}. CRIE-Agent first resolves the revision history into the final editing requirements, then plans their visual execution. The recovered intent guides both planning and verification, while reference-first grouping generates referenced objects before those that depend on them. This separates deciding which edits remain valid from organizing how their interdependent requirements are executed.

\paragraph{Visual Evaluation and Feedback.}
Question-based evaluation decomposes image requirements into fine-grained checks, as in TIFA and Davidsonian Scene Graph~\cite{hu2023tifa,cho2024dsg}. VIEScore distinguishes semantic consistency from perceptual quality in conditional image synthesis~\cite{ku2024viescore}, while TS2 examines how reliably evaluation metrics distinguish semantic errors~\cite{saxon2024ts2}. Visual feedback also supports iterative refinement through VQA-based assessment~\cite{singh2023divide} and repeated understanding, editing, and verification~\cite{bai2026uniediti}. In CRIE-Agent, verification checks complete candidates against the recovered final intent and guides retries and candidate selection. Each retry restarts the editing plan from the source image, allowing another attempt without carrying forward the image produced by a failed attempt.

\section{Implementation and Evaluation Details}
\label{sec:implementation_details}

\subsection{Dataset Split}
\label{sec:dataset_split}
We generate the corpus with a fixed random seed and shuffle it with a separate seed before dividing it into training, validation, and test sets. The splits share templates, vocabulary, and combinations of retained object categories. These category combinations specify object types, while each final-intent record also specifies their attributes and relations. Exact instructions and normalized final-intent records have no duplicates within or across splits.

\subsection{Model Settings}
\label{sec:model_settings}

\paragraph{Intent Agent training.}
The Intent Agent uses Qwen3-VL-8B-Instruct~\cite{qwen3vl2025}, fine-tuned on the 9,000 CRIE-S training instructions. The input consists of a system prompt and the editing instruction, without the source image. The target is the annotated final intent serialized as a JSON array. We freeze the pretrained weights and train LoRA adapters~\cite{lora} with rank 16, $\alpha=32$, and dropout 0.05, applied to the query, key, value, and output projections and the gate, up, and down projections. The loss is computed over the target JSON and end-of-sequence tokens, excluding the input prompt and padding.

We train for one epoch with BF16 computation on a single GPU, using a batch size of 4 and four gradient accumulation steps, giving an effective batch size of 16. The learning rate is $10^{-4}$, with a cosine schedule and a warmup ratio of 0.03. The maximum combined input and target length is 1,024 tokens.

\begin{table*}[t]
\centering
\caption{Single-step image editing on 50 naturally rephrased instructions. Scores are percentages, reported as mean $\pm$ sample standard deviation over three seeds. Bold values indicate the highest mean in each column.}
\label{tab:ood_natural_editing}
\begingroup
\small
\setlength{\tabcolsep}{5pt}
\renewcommand{\arraystretch}{1.08}
\begin{tabular}{@{}lccccc@{}}
\toprule
Intent recovery & Object $\uparrow$ & Attribute $\uparrow$ & Relations $\uparrow$ & Success $\uparrow$ \\
\midrule
VLM reorganization & $72.7\pm2.9$ & $76.6\pm1.7$ & $59.0\pm3.0$ & $2.7\pm1.2$ \\
\rowcolor{crieblue}
\textbf{Intent Agent} & $\mathbf{88.5\pm1.0}$ & $\mathbf{85.9\pm1.9}$ & $\mathbf{66.3\pm3.8}$ & $\mathbf{12.0\pm2.0}$ \\
\bottomrule
\end{tabular}
\endgroup
\end{table*}

\begin{table*}[t]
\centering
\caption{Generalization of final-intent recovery to different object counts and revision-history lengths. The reference setting contains five final objects and 10--12 editing operations. Accuracy measures complete recovery of the final intent.}
\label{tab:ood_count_history}
\small
\setlength{\tabcolsep}{6pt}
\begin{tabular}{@{}lcccc@{}}
\toprule
Test group & Final objects & History length & VLM reorganization $\uparrow$ & Intent Agent $\uparrow$ \\

Reference setting & 5 & 10--12 & 25.6--40.8 & 100.0 \\

\midrule

Object count & 3 & 8--10 & 39.2--60.0 & 100.0 \\
Object count & 4 & 9--11 & 36.8--50.4 & 100.0 \\

Object count & 6 & 11--13 & 28.0--47.2 & 100.0 \\
Object count & 7 & 12--14 & 28.0--50.4 & 99.2 \\
\midrule
Longer history & 5 & 16 & 14.4--28.8 & 92.0 \\
Longer history & 5 & 24 & 16.8--26.4 & 84.0 \\
Longer history & 5 & 32 & 14.4--28.0 & 84.0 \\
\bottomrule
\end{tabular}
\end{table*}

\paragraph{Output format.}
Each JSON entry describes one retained object and its final requirements. The \texttt{object} field specifies the object category, while optional fields record its color, pose, clothing, position, or relative size. Position and size descriptions refer to other objects by name. Each entry uses \texttt{"op": "add"} to express a final object requirement; the array does not reproduce the original revision history.

The following training example excludes a withdrawn rabbit, retains the flower's revised yellow color, and records the backpack's final size relative to the flower.

\begingroup
\small
\begin{verbatim}
[
  {
    "op": "add", "object": "dog",
    "color": "yellow", "pose": "stand still"
  },
  {
    "op": "add", "object": "man",
    "position": "to the right of the dog",
    "clothing": "shirt",
    "clothing_color": "blue"
  },
  {
    "op": "add", "object": "straw hat",
    "position": "to the left of the dog",
    "color": "yellow"
  },
  {
    "op": "add", "object": "flower",
    "position": "to the right of the man",
    "color": "yellow"
  },
  {
    "op": "add", "object": "backpack",
    "position": "to the left of the dog",
    "color": "blue",
    "size": "half as big as the flower"
  }
]
\end{verbatim}
\endgroup

\paragraph{Input template and decoding.}
Training and inference use the same system instruction:
\begin{quote}
\small
You turn a user's messy, self-correcting image-editing request into the minimal ordered list of edit operations. Collapse intermediate states, drop operations whose net effect is nothing, and keep every final attribute. Output ONLY a JSON list.
\end{quote}
At inference, we disable thinking and use greedy decoding with a maximum of 400 new tokens, stopping at an end-of-turn or end-of-text token. The generated text is parsed as JSON without constrained decoding.

\paragraph{Image editing.}
All five base editors are loaded in BF16, and the saved output images have a resolution of $1024\times1024$. Qwen-Image-Edit-2511~\cite{qwenimage} uses 30 denoising steps, while Qwen-Image-2.1~\cite{qwen2026image21}, FireRed-Image-Edit-1.1~\cite{firered-1.0}, and JoyAI-Image-Edit~\cite{song2026joyai} use 40 steps. FLUX.2-klein-4B~\cite{flux2} uses 4 steps with a guidance scale of 1. Classifier-free guidance is disabled for the two Qwen editors and uses a scale of 4 for FireRed and JoyAI. Unless otherwise specified, reference-first execution uses $K=2$ and permits at most three complete attempts. Each retry starts from the source image with a new editing seed and follows the same block plan.

\paragraph{External editing agents.}
ReasonEdit-S~\cite{yin2026reasonedit} uses Step1X-v1p2 with 50 denoising steps, a true classifier-free guidance scale of 6, and at most three diffusion calls. IntentEdit~\cite{zhang2026intentedit} uses InternVL3-14B~\cite{zhu2025internvl3} for instruction processing, with FLUX.1-Kontext~\cite{flux1} configured for 20 steps and a guidance scale of 3.5, and BAGEL~\cite{deng2025bagel} for 35 timesteps with text and image guidance scales of 4 and 2. Its workflow permits up to three outer iterations and five VLM attempts per module. RePlan~\cite{RePlan} uses Qwen2.5-VL-7B~\cite{qwen25vl2025} for planning and Qwen-Image-Edit-2511 for rendering, with 40 denoising steps and a true classifier-free guidance scale of 4. It performs one planning-and-rendering attempt per instruction. These agents retain their respective workflows and editing budgets.

\begin{figure*}[t]
\centering
\includegraphics[width=0.9\textwidth]{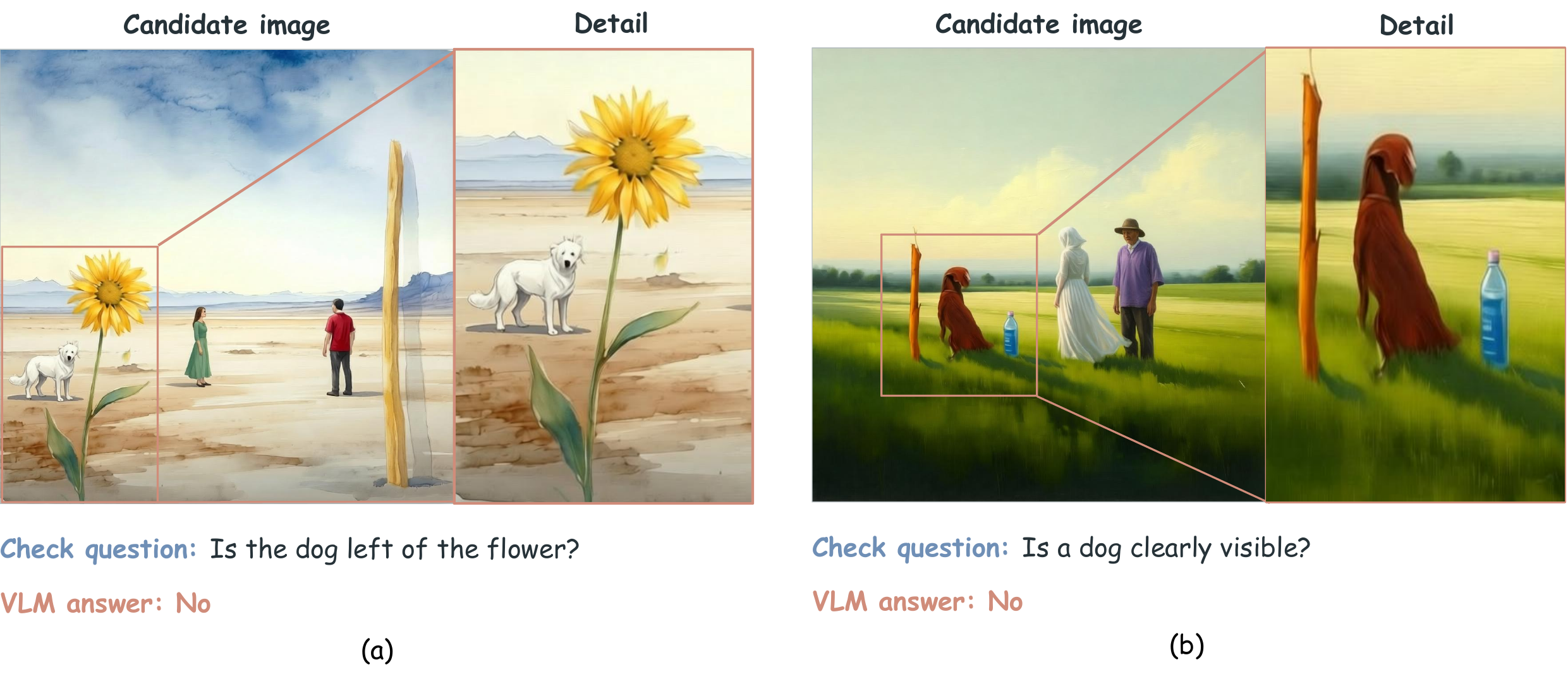}
\caption{Verification errors. The verifier rejects a satisfied left--right relation in (a) and overlooks a visible dog in (b), triggering a layout-guided retry and a retry without Layout, respectively.}
\label{fig:vlm_checks}
\end{figure*}

\subsection{Verification Details}
\label{sec:verification_details}

\paragraph{Visual checks.}
The verifier uses Qwen3-VL-8B-Instruct~\cite{qwen3vl2025} to answer yes/no questions about a candidate image. At the first answer-token position, let $z_t$ denote the logit of token $t$, and let $Y$ and $N$ contain the supported single-token variants of yes and no. We compute
\begin{equation}
 p_{\mathrm{yes}}=
 \frac{\sum_{t\in Y}\exp(z_t)}
 {\sum_{t\in Y}\exp(z_t)+\sum_{t\in N}\exp(z_t)}.
\label{eq:verification_probability}
\end{equation}
Fixed category-specific thresholds convert these scores into check outcomes. The question determines whether a positive answer indicates a satisfied requirement or an error. The resulting error report guides retries and candidate selection.

\paragraph{Retry and selection.}
The verifier checks a complete candidate for missing and extra objects, attribute errors, position errors, and deformations. Missing objects, extra objects, or incorrect position relations trigger another complete attempt when the retry budget permits. Layout supplies region hints on a retry when the preceding candidate contains a position error but no missing object and layout hints are available. Attribute and deformation errors affect candidate selection but do not independently trigger a retry.

Let $n_{\mathrm{obj}}$ denote the total number of missing and extra objects, and let $n_{\mathrm{attr}}$, $n_{\mathrm{pos}}$, and $n_{\mathrm{def}}$ denote attribute, position, and deformation errors. Artifact-free quality $s_{\mathrm{art}}$ and aesthetics $s_{\mathrm{aes}}$ take values from 1 to 5, with higher values better. The returned candidate minimizes the following key in lexicographic order:
\begin{equation}
\begin{aligned}
\kappa(r)=\bigl(&n_{\mathrm{obj}},\,
 n_{\mathrm{attr}}+n_{\mathrm{pos}},\,
 n_{\mathrm{def}},\\
 &5-s_{\mathrm{art}},\,5-s_{\mathrm{aes}}\bigr).
\end{aligned}
\label{eq:appendix_candidate_selection}
\end{equation}
Thus, object errors take priority, followed by attribute and position errors, deformation, and visual quality. Ties favor the earlier attempt.

\begin{table}[t]
\centering
\caption{Editor calls for the scheduling and block-capacity settings in Tables~\ref{tab:traversal} and~\ref{tab:block_capacity} on the CRIE-S test set. Values are the mean and standard deviation over three seeds, including all editing attempts.}
\label{tab:appendix_calls}
\small
\setlength{\tabcolsep}{5pt}
\begin{tabular}{@{}lcc@{}}
\toprule
Scheduling & $K$ & Editor calls \\
\midrule
Random & 2 & $6.92\pm0.09$ \\
Left-to-right & 2 & $7.85\pm0.12$ \\
Right-to-left & 2 & $7.56\pm0.07$ \\
\midrule
Reference-first & 1 & $9.95\pm0.33$ \\
Reference-first & 2 & $7.54\pm0.12$ \\
Reference-first & 3 & $6.93\pm0.07$ \\
Reference-first & 4 & $6.89\pm0.07$ \\
\bottomrule
\end{tabular}
\end{table}

\paragraph{Online checks and offline evaluation.}
Online verification controls retries and candidate selection. Offline evaluation uses its own fixed yes/no decision rule against the annotated final intent. Complete success requires every required semantic check to be answered and satisfied. Relative-size requirements are included in attribute evaluation, while the relation score measures spatial relations.
% TODO: Include exact probability normalization and category-specific decision
% inequalities in the released evaluation configuration.

\paragraph{Layout hints.}
Layout orders objects using the recovered left--right relations and assigns them to horizontal regions. Objects with fewer required predecessors are placed earlier; ties are resolved by prioritizing objects with more required successors, followed by their order in the predicted records. Objects without horizontal relations are moved toward the center without violating the required ordering. For the five-object setting, the regions are far left, left, center, right, and far right. These region descriptions are added to the editing prompts while retaining the original relation requirements. Whether to include the hints is determined separately for each retry using the preceding candidate's verification report.

\subsection{Editing Cost}
\label{sec:editing_cost}

Reference-first scheduling improves instruction fulfillment without requiring more average editor calls than the two spatial-order baselines at $K=2$ (Table~\ref{tab:appendix_calls}). Smaller blocks require more calls, whereas larger blocks reduce the number of editing steps. We count every editor call across all attempts, including candidates that are not selected. These counts exclude intent recovery and VLM verification calls. The reported call counts measure realized editor usage under each method's retry and early-stopping rules.

\section{Generalization of Intent Recovery}
\label{sec:intent_generalization}

We conduct additional evaluations to examine whether the Intent Agent learns to resolve revisions beyond the fixed formats of CRIE-S. We test natural expressions, different object counts, and longer revision histories, comparing the Intent Agent with VLM reorganization. Both use Qwen3-VL-8B-Instruct: VLM reorganization produces a paragraph of final requirements, while the fine-tuned Intent Agent produces object records that are compiled into an editing prompt.

\subsection{Natural Instruction Expressions}
\label{sec:natural_expression_ood}

\paragraph{Data and evaluation.}
This test contains 50 instructions with more conversational revision expressions, varied sentence structures, and contextual references than the original templates. The source images, revision histories, and final requirements remain unchanged. Both methods use Qwen-Image-Edit-2511 for single-step editing without the Execution Agent, with the same three seed settings. We use the same image evaluator and metrics as in the main experiments. Table~\ref{tab:ood_natural_editing} reports the mean and sample standard deviation over the three seeds.

\paragraph{Results.}
The Intent Agent improves all four reported metrics under natural rephrasing, including a 9.3 percentage point gain in complete success. The largest improvement is in object correctness, suggesting that recovering which objects should remain helps the editor handle canceled or superseded requests. These results support the usefulness of learned intent recovery beyond the training templates.

\begin{table*}[t]
\centering
\caption{Comparison of block capacity under reference-first scheduling on CRIE-S. Scores are percentages, reported as mean $\pm$ sample standard deviation over three seeds. Bold values indicate the highest mean before rounding; exact ties are both bolded.}
\label{tab:block_capacity}
\begingroup

\renewcommand{\arraystretch}{1.08}
\begin{tabular}{@{}lcccccc@{}}
\toprule
Strategy & $K$ & Object $\uparrow$ & Attribute $\uparrow$ & Relations $\uparrow$ & Success $\uparrow$ \\
\midrule
Reference-first & 1 & $94.6\pm0.4$ & $\mathbf{92.3\pm0.7}$ & $85.8\pm0.7$ & $34.0\pm1.5$ \\
\rowcolor{crieblue}
\textbf{Reference-first (Ours)} & 2 & $\mathbf{94.9\pm0.6}$ & $92.2\pm0.2$ & $87.3\pm0.4$ &  $\mathbf{39.3\pm4.3}$ \\
Reference-first & 3 & $94.7\pm0.4$ & $92.2\pm0.2$ & $\mathbf{87.4\pm1.1}$ & $37.3\pm0.8$ \\
Reference-first & 4 & $94.7\pm0.4$ & $92.3\pm0.3$ & $\mathbf{87.4\pm1.1}$ & $37.3\pm1.6$ \\
\bottomrule
\end{tabular}
\endgroup
\end{table*}

\begin{table*}[t]
\centering
\small
\renewcommand{\arraystretch}{1.10}
\caption{Results on CompBench's official multi-object addition subset. Ours uses the Execution Agent without the trained Intent Agent. Bold values indicate the best result within each backbone group.}
\label{tab:compbench72}
\begin{tabular}{@{}lccccc@{}}
\toprule
& \multicolumn{2}{c}{Foreground alignment}
& \multicolumn{3}{c}{Background preservation} \\
\cmidrule(lr){2-3}\cmidrule(lr){4-6}
Method & LC-T $\uparrow$ & LC-I $\uparrow$
& PSNR $\uparrow$ & SSIM $\uparrow$ & LPIPS $\downarrow$ \\
\midrule
\multicolumn{6}{l}{\textit{Backbone: Qwen-Image-Edit-2511}} \\
Direct editing
& \textbf{22.4823} & \textbf{0.8660}
& 21.1281 & 0.8927 & 0.1047 \\
RePlan
& 21.7825 & 0.8650
& \textbf{27.2761} & \textbf{0.9472} & \textbf{0.0416} \\
\rowcolor{crieblue}
\textbf{Ours}
& 22.3868 & 0.8593
& 21.8650 & 0.9143 & 0.0896 \\
\midrule
\multicolumn{6}{l}{\textit{Backbone: Qwen-Image-2.1}} \\
Direct editing
& 22.0172 & 0.8450
& 21.6768 & 0.7466 & 0.0898 \\
\rowcolor{crieblue}
\textbf{Ours}
& \textbf{22.2716} & \textbf{0.8541}
& \textbf{22.7831} & \textbf{0.7959} & \textbf{0.0805} \\
\bottomrule
\end{tabular}
\end{table*}

\subsection{Object Counts and Revision Histories}
\label{sec:count_history_generalization}

\paragraph{Data and evaluation.}
We evaluate eight groups of 125 instructions. Five groups contain three to seven final objects, including a five-object control, while retaining the original vocabulary and templates. The other three groups extend the five-object control histories to 16, 24, or 32 editing operations through additional revisions that preserve their final requirements.

This evaluation measures whether the recovered text preserves all final objects, attributes, and relations without adding requirements. Intent Agent records are compared programmatically with the annotations, allowing equivalent relations and differences in record order. VLM paragraphs are assessed through a model-based semantic review against the same annotations. Predictions are marked correct, incorrect, or unresolved when their meaning cannot be determined reliably. Both methods use deterministic decoding and are evaluated once per instruction. For unresolved VLM predictions, Table~\ref{tab:ood_count_history} reports lower and upper accuracy bounds by treating them as incorrect and correct, respectively; these are not confidence intervals. 

\paragraph{Results.}
The Intent Agent recovers the final intent correctly for 499 of 500 instructions with object counts different from the five-object training setting. For histories containing 16, 24, and 32 operations, recovery accuracy is 92.0\%, 84.0\%, and 84.0\%, respectively. Its accuracy exceeds the upper bound reported for VLM reorganization in every group, supporting transfer to the tested object counts and longer revision histories.

\section{Additional Experiments}
\label{sec:additional_cases}

\subsection{Evaluation with GLM}
\label{sec:independent_judge}

We re-evaluate the outputs in Table~\ref{tab:main} with GLM-4.6V-Flash as an independent judge, using the same annotated final intents and evaluation categories. Table~\ref{tab:independent_judge} reports the results separately from the Qwen3-VL-32B-Instruct results in the main paper. Across all five backbones, CRIE-Agent improves object correctness, attribute accuracy, spatial relations, and complete success over direct editing, with complete-success gains of 48.0--63.0 percentage points. 

\begin{table*}[t]
\centering
\caption{Independent evaluation with GLM-4.6V-Flash on CRIE-S. Bold values indicate the better result within each backbone pair.}
\label{tab:independent_judge}
\begingroup
\small
\setlength{\tabcolsep}{7pt}
\renewcommand{\arraystretch}{1.08}
\begin{tabular}{@{}lcccc@{}}
\toprule
Method & Object $\uparrow$ & Attribute $\uparrow$ & Relations $\uparrow$ & Success $\uparrow$ \\
\midrule
ReasonEdit-S & 68.68 & 63.62 & 47.00 & 0.50 \\
IntentEdit & 80.71 & 74.78 & 53.50 & 3.50 \\
RePlan & 46.75 & 20.69 & 10.38 & 0.00 \\
\midrule
Qwen-Image-Edit-2511 & 72.91 & 77.22 & 70.13 & 0.50 \\
\rowcolor{crieblue}
\quad \textbf{+ CRIE-Agent (Ours)} & \textbf{96.92} & \textbf{95.06} & \textbf{92.63} & \textbf{54.50} \\
\midrule
Qwen-Image-2.1 & 75.71 & 89.52 & 80.38 & 0.00 \\
\rowcolor{crieblue}
\quad \textbf{+ CRIE-Agent (Ours)} & \textbf{98.21} & \textbf{96.17} & \textbf{92.25} & \textbf{60.00} \\
\midrule
FLUX.2-klein-4B & 72.80 & 82.20 & 74.88 & 0.00 \\
\rowcolor{crieblue}
\quad \textbf{+ CRIE-Agent (Ours)} & \textbf{94.48} & \textbf{94.82} & \textbf{90.75} & \textbf{48.00} \\
\midrule
FireRed-Image-Edit-1.1 & 71.81 & 88.20 & 82.38 & 0.50 \\
\rowcolor{crieblue}
\quad \textbf{+ CRIE-Agent (Ours)} & \textbf{96.36} & \textbf{96.99} & \textbf{95.25} & \textbf{56.00} \\
\midrule
JoyAI-Image-Edit & 75.40 & 89.42 & 85.63 & 0.50 \\
\rowcolor{crieblue}
\quad \textbf{+ CRIE-Agent (Ours)} & \textbf{97.72} & \textbf{95.16} & \textbf{92.38} & \textbf{63.50} \\
\bottomrule
\end{tabular}
\endgroup
\end{table*}

\begin{table*}[t]
\centering
\caption{Editing with recovered final intents on the CRIE-S test set, evaluated by Qwen3-VL-32B-Instruct. All methods receive the final intents recovered by the Intent Agent. Scores are percentages. Bold values indicate the highest score within each backbone group.}
\label{tab:recovered_intent}
\begingroup
\small
\setlength{\tabcolsep}{6pt}
\renewcommand{\arraystretch}{1.08}
\begin{tabular}{@{}lccccc@{}}
\toprule
Execution & Object $\uparrow$ & Attribute $\uparrow$ & Relations $\uparrow$ & Success $\uparrow$ & Editor calls $\downarrow$ \\
\midrule
\multicolumn{6}{l}{\textit{Backbone: Qwen-Image-Edit-2511}} \\
Single-step & 91.97 & 88.00 & 69.38 & 11.00 & 1.00 \\
Multi-step & 89.29 & 86.36 & 74.50 & 17.00 & 10.51 \\
\rowcolor{crieblue}
CRIE-Agent (Ours) & \textbf{94.63} & \textbf{92.01} & \textbf{87.50} & \textbf{39.50} & 7.46 \\
\midrule
\multicolumn{6}{l}{\textit{Backbone: Qwen-Image-2.1}} \\
Single-step & 93.67 & 90.23 & 67.00 & 14.50 & 1.00 \\
Multi-step & 94.20 & 91.82 & 79.63 & 27.00 & 10.51 \\
\rowcolor{crieblue}
CRIE-Agent (Ours) & \textbf{96.78} & \textbf{93.50} & \textbf{88.00} & \textbf{50.00} & 6.88 \\
\midrule
\multicolumn{6}{l}{\textit{Backbone: FLUX.2-klein-4B}} \\
Single-step & 84.17 & \textbf{92.58} & 72.63 & 6.50 & 1.00 \\
Multi-step & 80.30 & 81.39 & 78.13 & 8.50 & 10.51 \\
\rowcolor{crieblue}
CRIE-Agent (Ours) & \textbf{91.79} & 91.95 & \textbf{85.75} & \textbf{27.50} & 7.25 \\
\midrule
\multicolumn{6}{l}{\textit{External agents with recovered intents}} \\
ReasonEdit-S & 75.19 & 89.20 & 57.38 & 2.50 & 2.52 \\
IntentEdit & 85.14 & 83.51 & 50.38 & 3.50 & 1.78 \\
RePlan & 51.29 & 29.09 & 9.30 & 0.00 & 1.00 \\
\bottomrule
\end{tabular}
\endgroup
\end{table*}

\subsection{Editing with Recovered Intents}
\label{sec:recovered_intent_results}

To examine whether recovering the final intent is sufficient for complex image editing, we provide the same recovered final intents to different execution methods. As shown in Table~\ref{tab:recovered_intent}, we compare single-step editing, fine-grained multi-step editing, existing editing agents, and our Execution Agent. The fine-grained baseline decomposes the recovered requirements into individual editing steps, whereas our Execution Agent coordinates their execution using reference-first scheduling and verification.

Across the three image editors, CRIE-Agent improves complete success over fine-grained multi-step editing by 19.0--23.0 percentage points, while using 6.88--7.46 editor calls per sample instead of 10.51. These results show that coordinated execution improves the realization of recovered intents beyond simply increasing the number of editing steps.

\subsection{Intent Recovery with Stronger Baselines}
\label{sec:stronger_intent_baselines}

We compare the trained Intent Agent with zero-shot Qwen3-VL-8B and Qwen3-VL-32B using an explicit JSON schema on the 200 CRIE-S test instructions. To assess the effect of the output budget, we evaluate both base models with maximum output lengths of 400 and 1,024 tokens. As shown in Table~\ref{tab:intent_baselines}, the Intent Agent achieves 100.0\% under the existing field-matching protocol, while Qwen3-VL-32B achieves 63.5\% with the longer budget. Increasing the output budget reduces the number of outputs that reach the limit, while the trained Intent Agent maintains a substantial advantage. The 32B runs use the same model, inputs, and prompts, but are executed on different hardware configurations.

\begin{table}[t]
\centering
\caption{Intent recovery on the CRIE-S test set. Accuracy follows the existing field-matching evaluation protocol. Truncated denotes the number of outputs that reached the token limit. Base models use zero-shot prompts with an explicit JSON schema; the Intent Agent uses its trained LoRA and original prompt. All results are evaluated on 200 instructions.}
\label{tab:intent_baselines}
\small
\setlength{\tabcolsep}{5pt}
\begin{tabular}{@{}lccc@{}}
\toprule
Method & Max tokens & Accuracy $\uparrow$ & Truncated $\downarrow$ \\
\midrule
Qwen3-VL-8B & 400 & 14.0 & 25 \\
Qwen3-VL-8B & 1,024 & 14.0 & 9 \\
Qwen3-VL-32B & 400 & 62.5 & 15 \\
Qwen3-VL-32B & 1,024 & 63.5 & 9 \\
\rowcolor{crieblue}
Intent Agent (Ours) & 400 & \textbf{100.0} & \textbf{0} \\
\bottomrule
\end{tabular}
\end{table}

\subsection{Block Capacity}
\label{sec:block_capacity_results}

Table~\ref{tab:block_capacity} compares block capacities under
reference-first scheduling. Among the tested settings, $K=2$ achieves
the highest mean complete success of 39.3\%, compared with 34.0\% for
$K=1$ and 37.3\% for $K=3$ and $K=4$. We select $K=2$ as the default configuration for the remaining experiments.

\subsection{Evaluation on an External Benchmark}
\label{sec:compbench}
We evaluate the Execution Agent without the trained Intent Agent on CompBench's official multi-object addition subset, where each instruction requires adding two objects. As shown in Table~\ref{tab:compbench72}, our Execution Agent improves all five reported metrics over direct editing with Qwen-Image-2.1. With Qwen-Image-Edit-2511, it improves the three background-preservation metrics, while the two foreground-alignment scores are slightly lower than direct editing.

One possible reason is that each instruction in this subset adds only two objects, leaving less room for reference-first scheduling than in the five-object CRIE-S setting. When the two objects can be edited jointly, decomposing the task may provide less benefit, while additional editing attempts and candidate selection can still affect the final result.

\subsection{Intent recovery and visual execution.}
\begin{figure*}[t]
\centering
\includegraphics[width=0.76\textwidth]{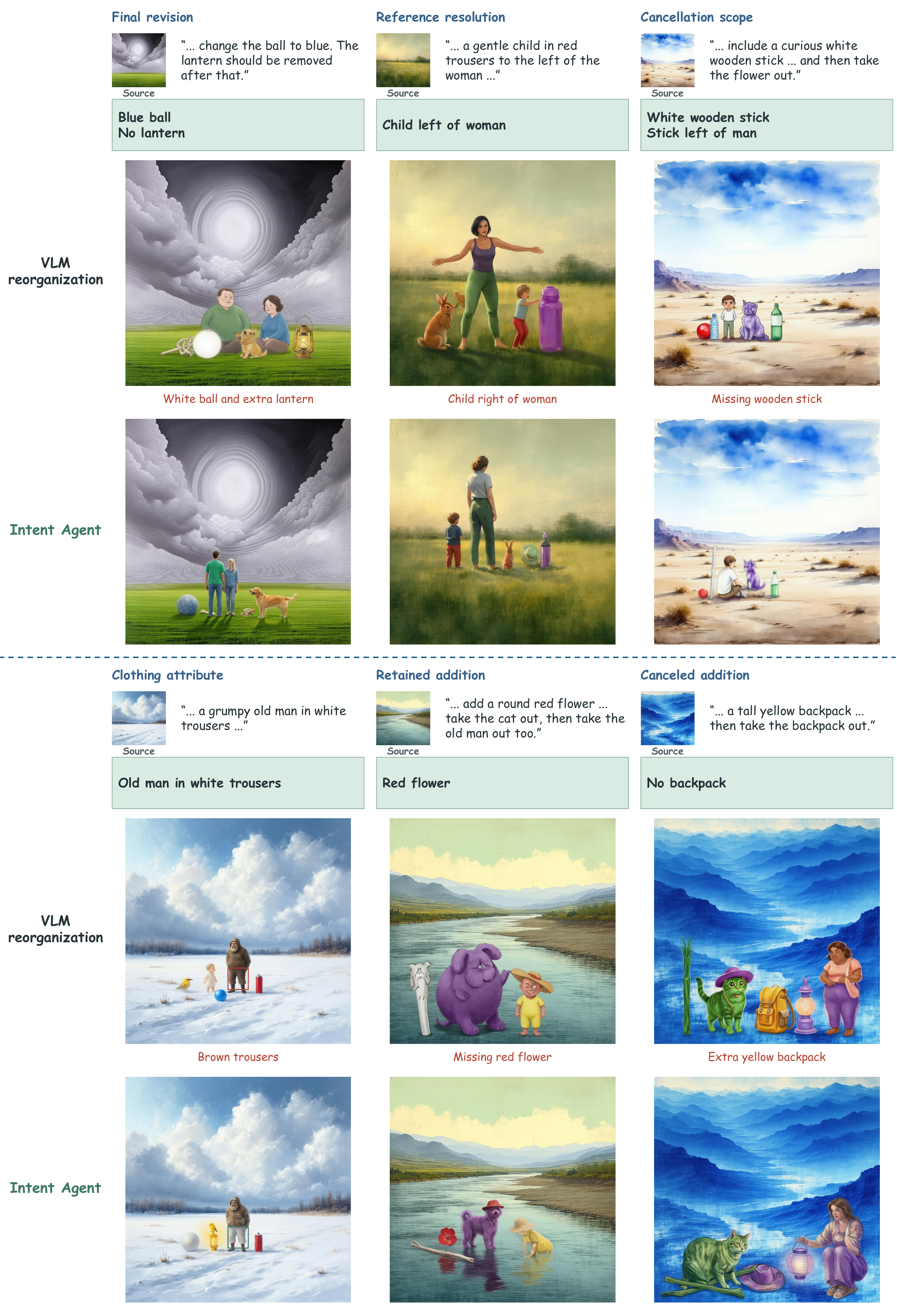}
\caption{Qualitative comparison on naturally expressed instructions. Both methods use single-step Qwen-Image-Edit-2511 editing with the same source image and editing seed for each pair. Instruction excerpts and key requirements are shown above the outputs. Red annotations indicate errors in VLM reorganization outputs.}
\label{fig:ood}
\end{figure*}

Figure~\ref{fig:ood} shows how the Intent Agent preserves the highlighted final requirements under natural expressions, including revised attributes, retained objects, and spatial references. The paired VLM-reorganization outputs exhibit errors in these requirements. The examples illustrate how resolving revisions before editing helps realize the intended objects and relations.

\subsection{Verification errors.}
Figure~\ref{fig:vlm_checks} shows two errors in visual verification. The verifier rejects a satisfied spatial relation in one example and overlooks a visible dog in the other, triggering a layout-guided retry and a retry without Layout, respectively. Such errors can cause unnecessary re-execution and affect candidate selection.

\end{document}